\documentclass[10pt,twocolumn,letterpaper]{article}

\usepackage{wacv}
\usepackage{times}
\usepackage{epsfig}
\usepackage{graphicx}
\usepackage{amsmath}
\usepackage{amssymb}
\usepackage{booktabs}
\usepackage[accsupp]{axessibility}

\usepackage{subcaption}
\usepackage{xcolor}

\newlength\savewidth\newcommand\shline{\noalign{\global\savewidth\arrayrulewidth
  \global\arrayrulewidth 1pt}\hline\noalign{\global\arrayrulewidth\savewidth}}
  
\newcommand{\beginsupplement}{%
        \setcounter{table}{0}
        \renewcommand{\thetable}{S\arabic{table}}%
        
        \setcounter{figure}{0}
        \renewcommand{\thefigure}{S\arabic{figure}}%
    
        \setcounter{equation}{0}
        \renewcommand{\theequation}{S\arabic{equation}}%
        
        \setcounter{section}{0}
        \renewcommand{\thesection}{S\arabic{section}}%
        
        \setcounter{page}{1}
        \renewcommand{\thepage}{S\arabic{page}}%
}

\def\wacvPaperID{1133} 

\wacvalgorithmstrack   

\wacvfinalcopy 

\ifwacvfinal
\usepackage[breaklinks=true,bookmarks=false]{hyperref}
\else
\usepackage[pagebackref=true,breaklinks=true,colorlinks,bookmarks=false]{hyperref}
\fi

\begin{document}

\title{Semantic Segmentation with Active Semi-Supervised Learning}

\author{Aneesh Rangnekar, Christopher Kanan, Matthew Hoffman\\
Rochester Institute of Technology\\
Rochester, NY, USA \\
{\tt\small aneesh.rangnekar@mail.rit.edu}
}

\maketitle
\thispagestyle{empty}

\begin{abstract}
   Using deep learning, we now have the ability to create exceptionally good semantic segmentation systems; however, collecting the prerequisite pixel-wise annotations for training images remains expensive and time-consuming. Therefore, it would be ideal to minimize the number of human annotations needed when creating a new dataset. Here, we address this problem by proposing a novel algorithm that combines active learning and semi-supervised learning. Active learning is an approach for identifying the best unlabeled samples to annotate. While there has been work on active learning for segmentation, most methods require annotating all pixel objects in each image, rather than only the most informative regions. We argue that this is inefficient. Instead, our active learning approach aims to minimize the number of annotations per-image. Our method is enriched with semi-supervised learning, where we use pseudo labels generated with a teacher-student framework to identify image regions that help disambiguate confused classes. We also integrate mechanisms that enable better performance on imbalanced label distributions, which have not been studied previously for active learning in semantic segmentation. In experiments on the CamVid and CityScapes datasets, our method obtains over 95\% of the network's performance on the full-training set using less than 17\% of the training data, whereas the previous state of the art required 40\% of the training data.
\end{abstract}

\section{Introduction}
\label{sec:intro}

Given enough labeled data, incredibly good semantic segmentation systems can be trained using deep learning~\cite{ronneberger2015u,badrinarayanan2017segnet,kohl2018probabilistic,kemker2018algorithms,chaudhary2019ritnet,zhu2019improving,rangnekar2020aerorit,tao2020hierarchical}. However, obtaining pixel-wise labels for semantic segmentation is incredibly time-consuming and expensive. For the COCO dataset, this required over 85,000 annotator hours\footnote{There were 10K hours for determining the categories present in each image, 20K for using point annotations for each object present, and over 55K for creating segmentation masks~\cite{lin2014microsoft}.}\cite{lin2014microsoft}. Thus, minimizing the number of annotations is desirable when creating a new semantic segmentation dataset. Active learning (AL) can help achieve this goal. AL is a framework for identifying the most informative samples in an unlabeled pool of samples for annotation. It has been heavily studied in computer vision for classification \cite{gal2017deep,beluch2018power,sener2018active,kirsch2019batchbald,yoo2019learning,Liu_2021_ICCV}, segmentation \cite{mackowiak2018cereals,sinha2019variational,kasarla2019region,Casanova2020Reinforced,colling2020metabox,li2020uncertainty,ebrahimi2020minimax,equal2020,xie2020deal,rangnekar2020semseghyperspectral,siddiqui2020viewal,Cai_2021_CVPR,Shin_2021_ICCV} and detection \cite{kao2018localization,roy2018deep,desai2019adaptive,aghdam2019active,harakeh2020bayesod,haussmann2020scalable,Yuan_2021_CVPR,Peng_2021_ICCV,Choi_2021_ICCV,elezi2021reducing}.

In AL, a machine learning system is iteratively retrained, where for each active learning cycle $C$ it 1) scores each unlabeled sample based on their informativeness, 2) requests annotations for $B$ samples, where $B$ is its annotation ``budget,'' and then 3) re-trains the system with the new samples. This loop continues until the desired performance is achieved. For semantic segmentation, most AL research has aimed to minimize the number of exhaustively annotated images \cite{mackowiak2018cereals,sinha2019variational,kasarla2019region,ebrahimi2020minimax,xie2020deal,siddiqui2020viewal}. For segmentation, we argue that this is sub-optimal. Rather than assuming images are exhaustively manually annotated, our approach aims to minimize the total number of manual annotations.  
\begin{figure}[t]
\begin{center}
\includegraphics[width=\linewidth]{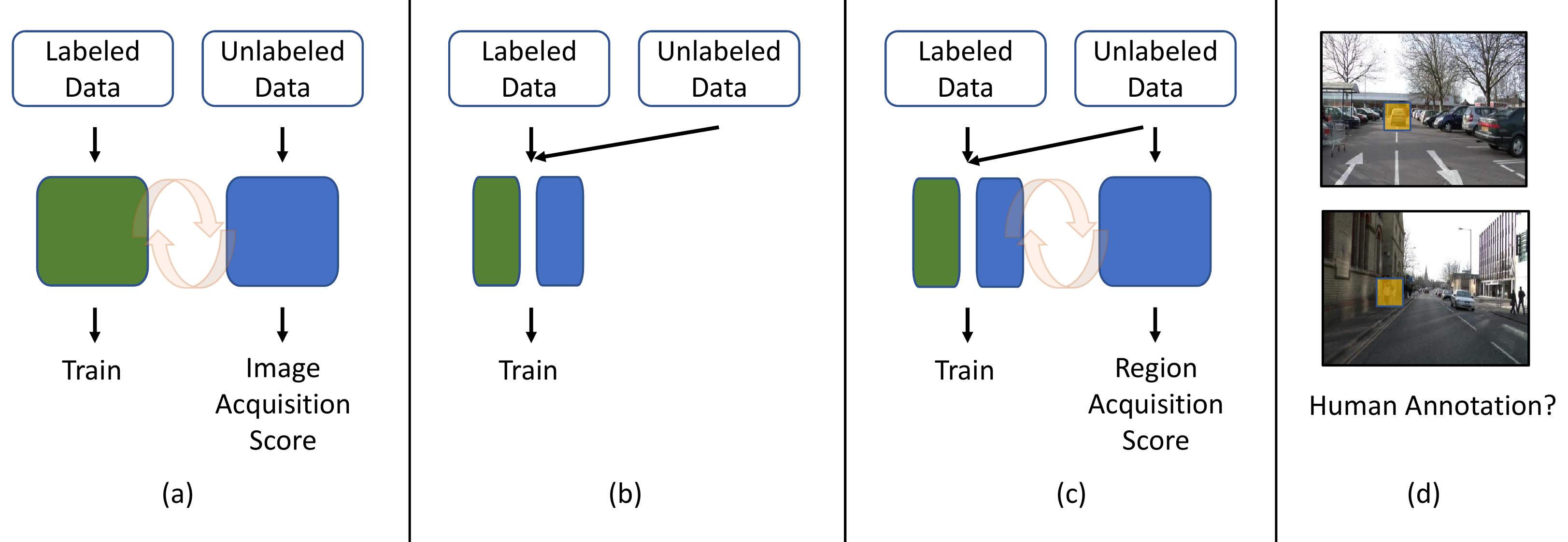}
\end{center}
\caption{
Our active learning approach aims to minimize the number of manual region annotations by using pseudo labels for unlabeled data during training phase. Blending (a) the traditional active learning approach with (b) the traditional semi-supervised approach, (c) our approach used semi-supervised learning on the unlabeled data to generate (d) region acquisition scores that can be queued for human annotations.}
\label{fig:teaser_frontpage}
\end{figure}

In our framework, only some fully-annotated images are provided manually, as the initial labeled set. The remaining annotations, for the unlabeled set, are produced automatically by employing semi-supervised learning (SSL) in each active learning cycle. Specifically, we employ pseudo labeling with a teacher-student framework to obtain the missing annotations. While pseudo labeling has been widely used for other problems~\cite{antti2017meanteacher,sohn2020fixmatch,Ouali_2020_CVPR,french2020semisupervised,liu2021bootstrapping,Chen_2021_CVPR,alonso2021semi}, it has not been used for active learning in semantic segmentation. We also introduce two regularization mechanisms for handling label imbalance. 

\paragraph{The key contributions of this paper are the following:}
\begin{itemize}
    \item We pioneer the application of SSL for active learning in semantic segmentation with a teacher-student framework,
    \item We show that two regularization schemes, Confidence Weighting and Balanced Classmix, improve generalization by mitigating dataset bias,
    \item Our system achieves performance that rivals or exceeds existing methods on the CamVid~\cite{brostow2009semantic} and CityScapes~\cite{cordts2016cityscapes} datasets.
\end{itemize}

\section{Related Work}
\label{sec:related}

Existing methods in active learning can be broadly classified in three categories, stream-learning \cite{vzliobaite2013active,santoro2016meta,wassermann2019ral}, query-synthesizing \cite{zhu2017generative,mahapatra2018efficient,liu2019generative,mayer2020adversarial} and query-computing. Stream-learning methods sequentially receive unlabeled samples and decide on the spot to request a label or discard. At the same time, query-synthesizing approaches often involve the use of generative adversarial networks (GANs) to ``synthesize" informative samples from the unlabeled data pool \cite{goodfellow2014generative}. Query-computing, or pool-based active learning, is the most common approach among the three and involves designing sample acquisition metrics for ranking the most informative samples.

Query-computing methods are further classified into applying Bayesian inference \cite{houlsby2011bayesian,gal2017deep,kirsch2019batchbald,feng2019lidaractive,li2020uncertainty,siddiqui2020viewal,harakeh2020bayesod}, ensemble learning \cite{beluch2018power,roy2018deep,haussmann2020scalable,rangnekar2020semseghyperspectral,Choi_2021_ICCV}, diversity learning \cite{sener2018active,yoo2019learning,desai2019adaptive,sinha2019variational,ebrahimi2020minimax,Liu_2021_ICCV,Cai_2021_CVPR} and representation learning \cite{mackowiak2018cereals,kao2018localization,aghdam2019active,kasarla2019region,Casanova2020Reinforced,colling2020metabox,gao2020consistency,equal2020,xie2020deal,Yuan_2021_CVPR,Shin_2021_ICCV,Peng_2021_ICCV,elezi2021reducing}. Most approaches use least confidence \cite{settles2009active}, softmax entropy \cite{shannon2001mathematical}, softmax margin \cite{scheffer2001active}, mutual information, Core-set \cite{sener2018active}, Monte-Carlo Dropout \cite{gal2016dropout}, or calculating the gradients of the output layer \cite{ash2019deep}. The premise for all approaches is the hypothesis that only the informative samples will stand out - for example, a low softmax margin indicates high confusion between two classes and hence is a high priority sample within the unlabeled data pool for labeling.

\textbf{Active Learning for Semantic Segmentation} spans all of the above venues and are addressed at image, region and pixel level. The Variational Adversarial Active Learning (VAAL) approach used adversarial learning to identify samples that confuse a learned variational auto-encoder and discriminator on whether the latent space indicates labeled or unlabeled data \cite{kingma2013auto,sinha2019variational}. The Minimax Active Learning (MAL) framework also used a discriminator to classify the most diverse samples as compared to the labeled set and paired it with class prototypes to identify the highest entropy samples \cite{ebrahimi2020minimax}. The Difficulty-awarE Active Learning (DEAL) architecture appended a probability attention branch to the standard semantic segmentation framework to learn to attend pixels belonging to the same semantic category before calculating metrics for acquisition \cite{xie2020deal}. Our approach is encouraged from VAAL and MAL, wherein we improve on the idea of using unlabeled data alongside labeled data, but with pseudo labels instead of an adversarial framework (Sec. \ref{subsec:mainresults}).

On a region level, \cite{mackowiak2018cereals} introduced regional Monte-Carlo Dropout with spatial diversity and cost analysis to select regions within images for labeling while RALIS used reinforcement learning to determine the best blocks to sample  \cite{Casanova2020Reinforced}. ViewAL used diversity in scene object views in multi-view dataset samples \cite{siddiqui2020viewal}, \cite{kasarla2019region} refined labels with Conditional Random Fields, and the paper \cite{Cai_2021_CVPR} introduced a class-balanced sampling to select super-pixels regions generated by the SEEDS algorithm \cite{van2012seeds}. Recently, PixelPick reduced the labeling costs by a significant degree by training networks only with sparse pixel annotations \cite{Shin_2021_ICCV}. However, most of these approaches fail to utilize the unlabeled data for training the models to their maximum potential. 

A notable exception is EquAL, an active learning approach that incorporates self-consistency on the image and its horizontally flipped version \cite{equal2020}. The authors then use the same constraint as the acquisition metric for queuing regions to be labeled within images of the unlabeled pool. It is worth mentioning that similar ideas have been successful in reducing data labeling costs for classification \cite{gao2020consistency} and object detection \cite{elezi2021reducing}. We further develop this line of work to use semi-supervised learning for improving active learning frameworks. More specifically, instead of using consistency in terms of equivariance and data augmentations, we propose to work on pseudo labeling to leverage the unlabeled data pool to its maximum potential.

\textbf{Semi-Supervised Learning for Semantic Segmentation} is typically addressed in three ways, explicit consistency regularization \cite{zou2020pseudoseg,feng2020dmt,Ouali_2020_CVPR,Chen_2021_CVPR,Lai_2021_CVPR}, using a teacher-student framework \cite{mittal2019semi,french2020semisupervised,Olsson_2021_WACV}, and recently, combining teacher-student with contrastive embeddings \cite{liu2021bootstrapping,zhou2021c3}.

Teacher-student training pipelines use the mean-teacher framework that maintains an exponentially moving averaged copy of the student model to provide smoother pseudo labels \cite{antti2017meanteacher}. This approach is trained by passing weak image copies to the teacher, acquiring pseudo labels and then training the student on strong perturbed copies of the same set of images  \cite{berthelot2019mixmatch,sohn2020fixmatch}. The paper \cite{mittal2019semi} used GAN based learning to encourage confusions between predictions from the labeled and unlabeled examples. CutMix-Seg showed that applying CutMix boosts the performance of learned representations for semantic segmentation \cite{yun2019cutmix,french2020semisupervised}, while ClassMix modified the CutMix mechanism to sample masks corresponding to individual classes for mixing \cite{Olsson_2021_WACV}. Recently, Regional Contrast (ReCo) and C$^{3}$-SemiSeg learned contrastive pixel-embeddings to further strengthen the representations alongside conventional cross-entropy as a separate dedicated branch \cite{liu2021bootstrapping,zhou2021c3}. For simplicity and to set a baseline, we adopt the mean-teacher framework to generate pseudo labels on the unlabeled data pool and expand on this idea for active learning in the next section.

\section{Our Approach}
\label{sec:methods}
\begin{figure*}[t]
\begin{center}
\includegraphics[width=0.95\linewidth]{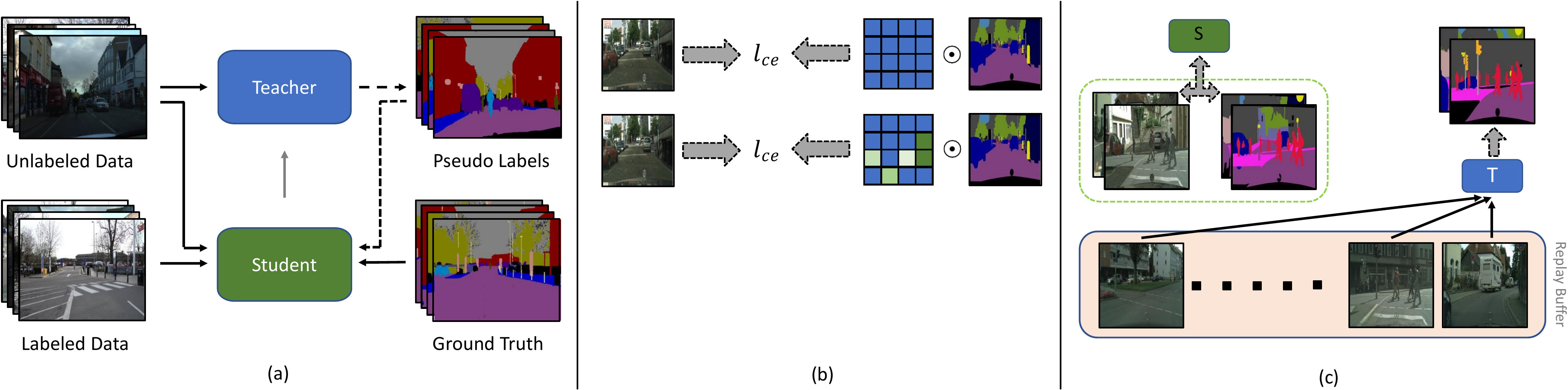}
\end{center}
\caption{
Our SSL setup with (a) teacher-student framework, where (b) shows Confidence Weighting on a per-pixel basis, where the overall cross-entropy loss is weighted by the highest probability for the corresponding pseudo label (blue indicates high, and green indicates low), and (c) shows the Balanced ClassMix, where we maintain a replay buffer of unlabeled images and randomly sample them for ClassMix to further boost the number of samples for tail classes.}
\label{fig:algooverviewdetail}
\end{figure*}

Active learning frameworks have a standard operational cycle - the network is trained on labeled data - for semantic segmentation, this is typically with cross-entropy loss, and then used to score samples within the unlabeled data pool (Fig. \ref{fig:teaser_frontpage}a). The learned network is used to infer statistics in terms of sample information within the unlabeled data pool, represented as scores for data acquisition. Every active learning cycle $C$, a portion of samples from the unlabeled pool - ``budget" $B$, is sampled, annotated and added to the labeled set. This loop continues till either the budget is exhausted or the updated labeled data pool achieves acceptable performance.

Active learning for semantic segmentation comes with three significant ordeals. Segmentation datasets have their own distribution bias - there is a set of classes that are heavily under-represented ``tail distribution", for example, column pole and sign in CamVid \cite{brostow2009semantic}, and rider and train in CityScapes \cite{cordts2016cityscapes}. They also have significant variance in the within-scene imagery, for example, illumination, contrast, and viewpoint changes. The most common semantic segmentation models use the ResNet or MobileNet family of backbones, with either Fully Convolutional Networks (FCNs), Dilated Residual Networks (DRNs) or DeepLab family of architectures coupled with additional blocks for inference \cite{He_2016_CVPR,Sandler_2018_CVPR,Howard_2019_ICCV,Long_2015_CVPR,Yu_2017_CVPR,Chen2018DeepLabSI,chen2017rethinking,Chen_2018_ECCV}. A common ingredient in all these networks is the Batch Normalization layer, which coupled with the dataset and scenery bias, makes standard training of neural networks difficult to generalize to new unseen data \cite{ioffe15}. This brings us to the third problem - in an active learning scenario, the unlabeled images are ranked in order of a network's response, which may be biased based upon how well it has learned on a relatively smaller training set that has its own distribution of classes.

To mitigate these challenges, we leverage unlabeled data using semi-supervised learning during each active learning cycle $C$ as a means of understanding the unlabeled data distribution (Fig. \ref{fig:teaser_frontpage}b). We use a teacher-student framework for generating pseudo labels on a per-pixel basis to learn better representations of the unlabeled data pool while assisting the network's learning with labeled data. We also replace sampling entire images with regions instead (similar to \cite{mackowiak2018cereals,Casanova2020Reinforced,equal2020,colling2020metabox}, Fig. \ref{fig:teaser_frontpage}c, d). 

On a higher level, we initialize two sets of the network, wherein the student network is trained with cross-entropy and the teacher network is updated in a gradual manner with the student's parameters (Fig. \ref{fig:algooverviewdetail}a, Eqn. \ref{eq:mt_ce}). The teacher network is then used to infer pseudo labels on the unlabeled images, which are used to train the student network on the unlabeled data with cross-entropy. At the end of the SSL training for each active learning cycle, we rank the regions within the unlabeled pool of images based on an acquisition metric computed with the teacher's performance and then select the highest ranking set for labeling. We keep retraining the dataset in an SSL fashion for multiple active learning cycle till a satisfactory performance is achieved to end the label acquisitions.

A slow moving update (teacher network) results in more stable cumulative predictions, however, simply adopting this approach is difficult due to the overall class distribution bias in the labeled data. This is undesirable as pseudo labels from the teacher act as ground-truths for the student network, which may quickly learn the false representations, especially for tail distribution classes. Furthermore, when regions from the unlabeled images are scored as candidates for labeling, a poorly trained network combination may provide a significant shift in the scores for regions that otherwise would be necessary for correcting the bias and assist in guiding the next active learning cycle. To circumvent this problem, we introduce two regularizing schemes - Confidence Weighting and Balanced Classmix.

\subsection{Teacher-Student Framework}
\label{subsec:meanteacherexplanation}
The mean teacher framework, proposed for semi-supervised learning on image classification, consists of two networks, a student ($\theta$) and a teacher ($\theta'$) \cite{antti2017meanteacher}. The student and teacher share the same network architecture, and the teacher is updated by an exponential moving average (EMA) of the student parameters (Eqn. \ref{eq:mt_update}):

\begin{equation}
    \theta':=m\theta'+(1-m)\theta,
    \label{eq:mt_update}
\end{equation}
where $m$ is the smoothness coefficient (momentum) and is set to $0.99$ \cite{french2020semisupervised,Olsson_2021_WACV}. For the labeled images, the student is learned using a supervised loss (cross-entropy) with the ground truth information (Eqn. \ref{eq:mt_ce}):

\begin{equation}
    \mathcal{L}_{sup} = \ell_{ce}(\theta(x_l),y_l),
    \label{eq:mt_ce}
\end{equation}
where $x_l$ are the inputs to the network and $y_l$ are the corresponding ground truth labels. The continuously updated teacher model (Eqn. \ref{eq:mt_update}) is used to generate pseudo labels on the unlabeled data which are used to train the student network. The teacher receives weakly-augmented version of the images ($x_{u-w}$) to generate predictions and the student model is trained using the generated pseudo labels as ground truths following (Eqn. \ref{eq:mt_unsup1}):

\begin{equation}
    \mathcal{L}_{unsup} = \ell_{ce}(\theta(x_{u-s}),[\theta'(x_{u-w})]),
    \label{eq:mt_unsup1}
\end{equation}
where $x_{u-s}$ is the strong-augmented (perturbed) version as input to the student. The $[]$ indicate the conversion of the logits to one hot vectors, which are used as ground truths for training. We use random flip and random crop operations for $x_{u-w}$, and random scaling, random flip, color jittering and ClassMix for $x_{u-s}$.

We treat $[\theta'(x_{u-w})]$ as a one-hot vector indicating the corresponding pseudo label for the pixel under consideration. The final loss for training is then formulated as:

\begin{equation}
    \mathcal{L}_{total} = \mathcal{L}_{sup} + \eta\cdot\mathcal{L}_{unsup},
    \label{eq:mt_total_stage1}
\end{equation}
where $\eta$ corresponds to the weighting for the unsupervised loss. We calculate $\eta$ as the ratio of the number of pixels within an image that satisfy $p$ $>$ 0.97, (wherein $p$ indicates the max probability for the pseudo label for the pixel) to the total number of pixels within that image, similar to \cite{french2020semisupervised,Olsson_2021_WACV,liu2021bootstrapping}. 
\subsection{Confidence Weighting}
\label{subsec:confidenceweightingexplanation}

Eqn. \ref{eq:mt_total_stage1} weighs the contribution of pseudo labeled pixels from the teacher network based on the amount of corresponding probabilities $p$ which pass a preset confidence threshold \cite{french2020semisupervised,Olsson_2021_WACV,liu2021bootstrapping}. However, the weight is applied on an image-level for a mini-batch setting, which implies that all pixels within an image are trained with the same importance irrespective of their prediction confidence. We hypothesize that the student network can quickly overfit to the distribution of classes present in the limited training set and as the teacher is a slow update (Eqn. \ref{eq:mt_update}), this can eventually introduce bias with respect to classes that have limited pixel annotations. 

Most semi-supervised learners function on a one-time run agenda. In an active learning setting, however, a class distribution bias can prove more harmful if the networks become very confident of their predictions and they have false confidence on their predictions when computing an acquisition metric. In a long run, this is undesirable as the entire goal of active learning is to continuously query informative samples, minimize labeling costs and yet maintain good performance.

There are multiple approaches in semi-supervised image classification which strive to solve this class imbalance problem, which are broadly based on re-sampling or re-weighting \cite{berthelot2019remixmatch,jamal2020rethinking,wei2021crest}. Most of the approaches involve keeping a running statistic of the number of images falling under a certain class label, which can be daunting when directly applied to semantic segmentation on a pixel basis as the head classes (sky, road, building) heavily overshadow the tail classes (bicycle, pedestrian, fence) . Following the fundamentals of these approaches, we propose a simple modification to Eqn. \ref{eq:mt_unsup1} as follows:

\begin{equation}
    \mathcal{L}_{unsup} = \ell_{ce}(\theta(x_{u-s}),p\cdot[\theta'(x_{u-w})]),
    \label{eq:mt_unsup2}
\end{equation}
where $p$ is the corresponding max probability for every pseudo label from the teacher network (Fig. \ref{fig:algooverviewdetail}b). This approach ensures that high confidence pseudo labels and annotated labels within an active learning cycles get more importance as compared low confidence pseudo labels.

\subsection{Balanced ClassMix}
\label{subsec:adaptiveclassmixexplanation}

Our second regularization scheme focuses on tail classes imbalance from an oversampling perspective with data augmentation. For this task, we build on ideas from continual learning and ClassMix \cite{rebuffi2017icarl,balaji2020effectiveness,hayes2020remind,hayes2021replay,Olsson_2021_WACV}. Replay buffers are widely used to mitigate catastrophic forgetting by storing previous sets of samples and mixing them with new samples for training neural networks \cite{nguyen2017variational,van2019three,hayes2020remind,hayes2021replay}. These buffers are usually restricted in  size due to memory constraints and there are multiple schemes for replay sampling, but, for our task, we rely on uniform sampling from the buffer.

Specifically, we initialize a replay buffer with limit $M$. At every iteration, we add images into the buffer for sampling later. For the current iteration, we randomly sample images of the same batch size from the replay buffer. These set of images are then passed through the same pipeline for calculating Eqn. \ref{eq:mt_unsup2}, except we modify the probability distribution for ClassMix data augmentation. ClassMix uniformly samples the mask from the entire set of classes in its original setting. For our setting, we bias the sampling rate to focus on more samples from the tail classes as compared to the head classes, which is assessed based on the distribution within the labeled data. We accomplish this by sampling classes for ClassMix by two separate distributions, head and tail, instead of a single combined distribution. This assists in the active learning loops as we expect the regions queued by the acquisition metric for labeling belong to the tail classes, due to more challenges posed during teacher-student training with our ClassMix variant (Fig. \ref{fig:algooverviewdetail}c). 

Thus, for the current mini-batch, we get two different $x_{u_s}$, where the first set corresponds to within mini-batch augmentations (${L}_{unsup1}$), and the second set corresponds to replay augmentations (${L}_{unsup2}$). For computation efficiency, we only maintain the ClassMix-d versions equal to the batch size. It is entirely plausible to use standard ClassMix with the replay images, however, our formulation ensures that more pixels from the tail classes are seen during training, especially post a few cycles of label acquisitions on the unlabeled data pool. Our total training loss from Eqn. \ref{eq:mt_total_stage1} becomes, 

\begin{equation}
    \mathcal{L}_{total} = \mathcal{L}_{sup} + \eta_{1}\cdot\mathcal{L}_{unsup1} + \eta_{2}\cdot\mathcal{L}_{unsup2},
    \label{eq:mt_total_stage2}
\end{equation}
where the values of $\eta$ in Eqn. \ref{eq:mt_total_stage2} are still calculated in the same manner as mentioned before.

\subsection{Sampling Strategy}
\label{subsec:summary}

Our entire framework, semi-supervised semantic segmentation for active learning (S4AL) is a two-step process where we follow the standard protocols of active learning and iterate over multiple active learning cycles $C$, but at the same time, use pseudo labeling to leverage the unlabeled data pool in a much efficient manner. A natural choice for sampling shifts from image-level to region-level as there are multiple sub-regions whose predictions become more confident over cycles, thus lowering the overall image scores and increasing the possibility of missing out on key annotations that may belong to tail classes. For our acquisition metric, we adopt four strategies: random sampling, least confidence, softmax entropy, and softmax margin. We refer the readers to \cite{siddiqui2020viewal} for an in-depth explanation for all sampling strategies in active learning. From our initial set of experiments, we found softmax entropy to be the best suited acquisition metric for all our datasets (details are enclosed in the supplemental).


\begin{table*}[t]
\tiny
\caption{\textbf{IoU:} Class-wise and mean on CityScapes using MobileNetv2 - while all approaches use 40\% data to achiever their goal, we achieve our goal with only 16\% data.}
\label{tab:results_cityscapes}
\def\arraystretch{1.2}
\resizebox{\textwidth}{!}{%
\begin{tabular}{lcccccccccc}
Method & Road & \begin{tabular}[c]{@{}c@{}}Side\\ walk\end{tabular} & Building & Wall & Fence & Pole & \begin{tabular}[c]{@{}c@{}}Traffic\\ Light\end{tabular} & \begin{tabular}[c]{@{}c@{}}Traffic\\ Sign\end{tabular} & Vegetation & Terrain \\ \shline
{\color[HTML]{9B9B9B} Supervised} & {\color[HTML]{9B9B9B} 97.58} & {\color[HTML]{9B9B9B} 80.55} & {\color[HTML]{9B9B9B} 88.43} & {\color[HTML]{9B9B9B} 51.22} & {\color[HTML]{9B9B9B} 47.61} & {\color[HTML]{9B9B9B} 35.19} & {\color[HTML]{9B9B9B} 42.19} & {\color[HTML]{9B9B9B} 56.79} & {\color[HTML]{9B9B9B} 89.41} & {\color[HTML]{9B9B9B} 60.22} \\
Random & 96.03 & 72.36 & 86.79 & 43.56 & 44.22 & 36.99 & 35.28 & 53.87 & 86.91 & 54.58 \\
Entropy & 96.28 & 73.31 & 87.13 & 43.82 & 43.87 & 38.10 & 37.74 & 55.39 & 87.52 & 53.68 \\
Core-Set \cite{sener2018active} & 96.12 & 72.76 & 87.03 & 44.86 & 45.86 & 35.84 & 34.81 & 53.07 & 87.18 & 53.49 \\
DEAL \cite{xie2020deal} & 95.89 & 71.69 & 87.09 & 45.61 & 44.94 & 38.29 & 36.51 & 55.47 & 87.53 & 56.90 \\
\textbf{S4AL} & 97.73 & 81.76 & 88.63 & 51.42 & 47.40 & 36.00 & 43.91 & 58.27 & 89.72 & 62.01 \\ \hline


 & Sky & \begin{tabular}[c]{@{}c@{}}Pedes-\\ trian\end{tabular} & Rider & Car & Truck & Bus & Train & \begin{tabular}[c]{@{}c@{}}Motor-\\ Cycle\end{tabular} & Bicycle & mIoU \\ \shline
{\color[HTML]{9B9B9B} Supervised} & {\color[HTML]{9B9B9B} 92.69} & {\color[HTML]{9B9B9B} 65.12} & {\color[HTML]{9B9B9B} 37.32} & {\color[HTML]{9B9B9B} 90.67} & {\color[HTML]{9B9B9B} 66.24} & {\color[HTML]{9B9B9B} 71.84} & {\color[HTML]{9B9B9B} 63.84} & {\color[HTML]{9B9B9B} 42.35} & {\color[HTML]{9B9B9B} 61.84} & {\color[HTML]{9B9B9B} 65.30} \\

Random & 91.47 & 62.74 & 37.51 & 88.05 & 56.64 & 61.00 & 43.69 & 30.58 & 55.67 & 59.00 \\
Entropy & 92.05 & 63.96 & 34.44 & 88.38 & 59.38 & 64.64 & 50.80 & 36.13 & 57.10 & 61.46 \\
Core-Set \cite{sener2018active} & 91.89 & 62.48 & 36.28 & 87.63 & 57.25 & 67.02 & 56.59 & 29.34 & 53.56 & 60.69 \\
DEAL \cite{xie2020deal} & 91.78 & 64.25 & 39.77 & 88.11 & 56.87 & 64.46 & 50.39 & 38.92 & 56.59 & 61.64 \\
\textbf{S4AL} & 92.81 & 65.62 & 39.71 & 90.52 & 66.07 & 65.31 & 46.03 & 46.88 & 61.77 & 64.80 \\ \hline
\end{tabular}%
}
\end{table*}
\begin{table*}[t]
    \caption{\textbf{Discussions:} of various studies conducted on CamVid and CityScapes dataset. We use MobileNetv2 for all our experiments, except in Table \ref{tab:comparisons_drn_approaches}.}
    \label{tab:compiled_results}
    \begin{subtable}[t]{0.30\textwidth}
        \def\arraystretch{1.2}
        \small
        \caption{\textbf{Region vs Image:} the entire dataset, with top three classes with the largest difference in IoU and (Recall).}
        \label{tab:regionvsimage_camvid}
        \centering
        \resizebox{\textwidth}{!}{%
        \begin{tabular}{@{}lll@{}}
         & \multicolumn{2}{c}{IoU} \\ 
         & Region & Image \\ \shline
        CamVid & 61.78 & 60.31 \\
        $\cdot$ Sign & 40.27 (73.75) & 34.66 (75.22) \\
        $\cdot$ Pedestrian & 49.96 (70.30) & 43.99 (68.29) \\
        $\cdot$ Bicyclist & 51.31 (77.14) & 49.76 (72.08) \\ \bottomrule
        \end{tabular}%
        }
        \end{subtable}
    \hfill
    \begin{subtable}[t]{0.30\textwidth}
        \def\arraystretch{1.2}  
        \small
        \caption{\textbf{Region vs Image:} the entire dataset, with top three classes with the largest difference in IoU and (Recall).}
        \label{tab:regionvsimage_cityscapes}
        \centering
        \resizebox{\textwidth}{!}{%
        \begin{tabular}{@{}lll@{}}
         & \multicolumn{2}{c}{IoU} \\ 
         & Region & Image \\ \shline
        CityScapes & 64.70 & 64.73 \\
        $\cdot$ Truck & 66.07 (85.17) & 60.12 (74.77) \\
        $\cdot$ Bus & 65.31 (89.05) & 67.13 (81.91) \\
        $\cdot$ Train & 46.03 (51.07) & 57.30 (66.67) \\ \bottomrule
        \end{tabular}%
        }
        
     \end{subtable}
     \hfill
     \begin{subtable}[t]{0.36\textwidth}
        \def\arraystretch{1.2}  
        \small
        \caption{\textbf{IoU:} using DRN on VAAL \cite{sinha2019variational} and MAL \cite{ebrahimi2020minimax}. IoU$_s$ indicates the score at starting with 10\% of the data, IoU$_f$ indicates the score at end of active learning, IoU$_{sup}$ are the scores with fully supervised data, and the last row indicates the amount of data utilized.}
        \label{tab:comparisons_drn_approaches}
        \centering
        \resizebox{\textwidth}{!}{%
        \begin{tabular}{@{}llll@{}}
        \\
        \multicolumn{1}{c}{} & VAAL \cite{sinha2019variational} & MAL \cite{ebrahimi2020minimax} & \textbf{S4AL} \\ \shline
        IoU$_s$ & 46.2 & 48.9 & 57.9 \\
        IoU$_f$ & 56.5 {\color[HTML]{32CB00}($+$10.3)} & 58.4 {\color[HTML]{32CB00}($+$9.5)} & 65.7 {\color[HTML]{32CB00}($+$7.8)} \\
        IoU$_{sup}$ & 62.95 $\pm$ 0.70 & 61.9 $\pm$ 0.70 & 67.68 $\pm$ 1.2 \\
        \% data & 40 & 30 & 16 \\ \bottomrule
        \end{tabular}%
        }
        
    \end{subtable}

\end{table*}
\begin{figure*}
    \begin{subfigure}[t]{0.48\textwidth}
    \begin{center}
    \includegraphics[width=\linewidth]{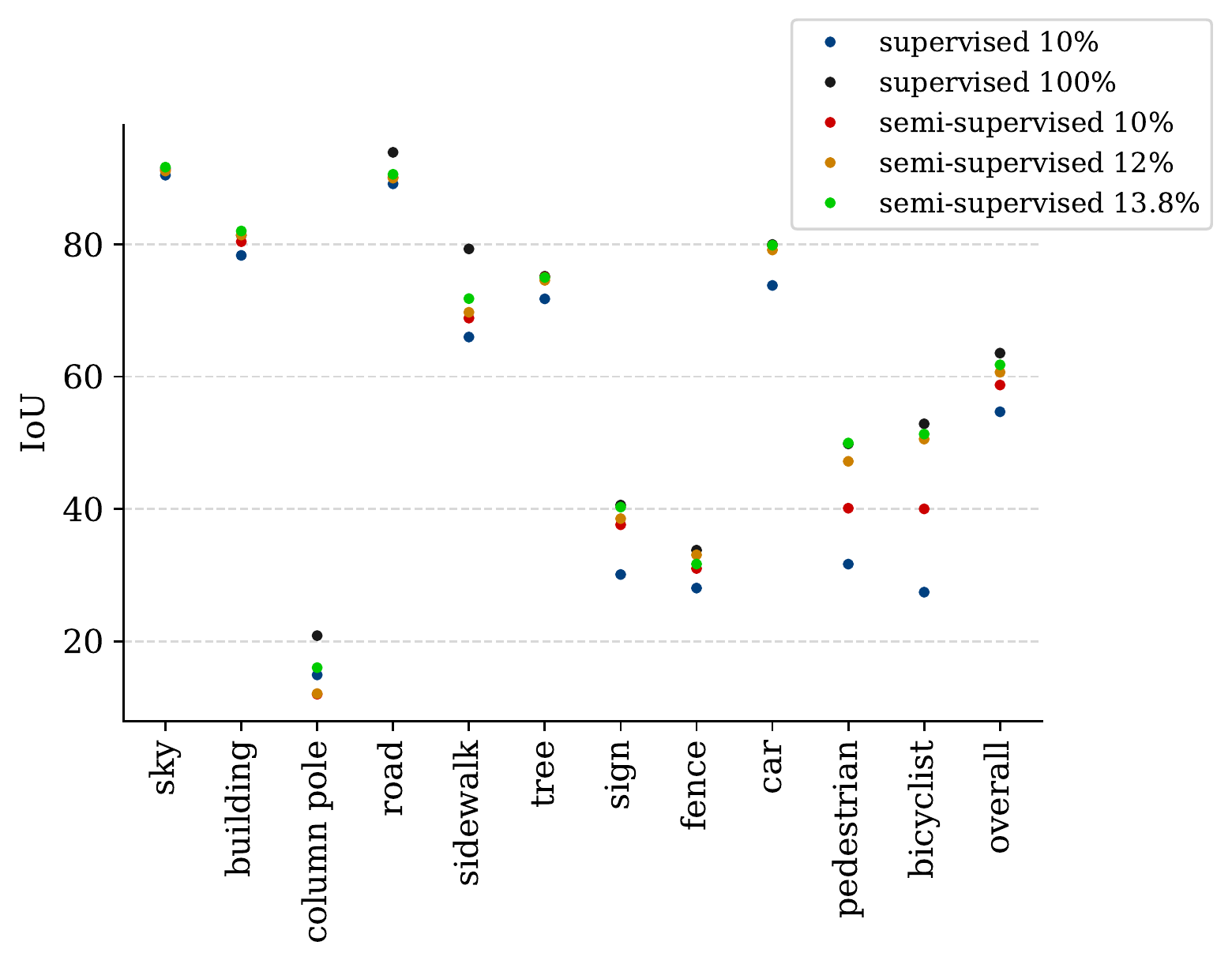}
    \end{center}
    \caption{CamVid}
    \label{fig:results_camvid_main}
    \end{subfigure}
    \begin{subfigure}[t]{0.48\textwidth}
    \begin{center}
    \includegraphics[width=\linewidth]{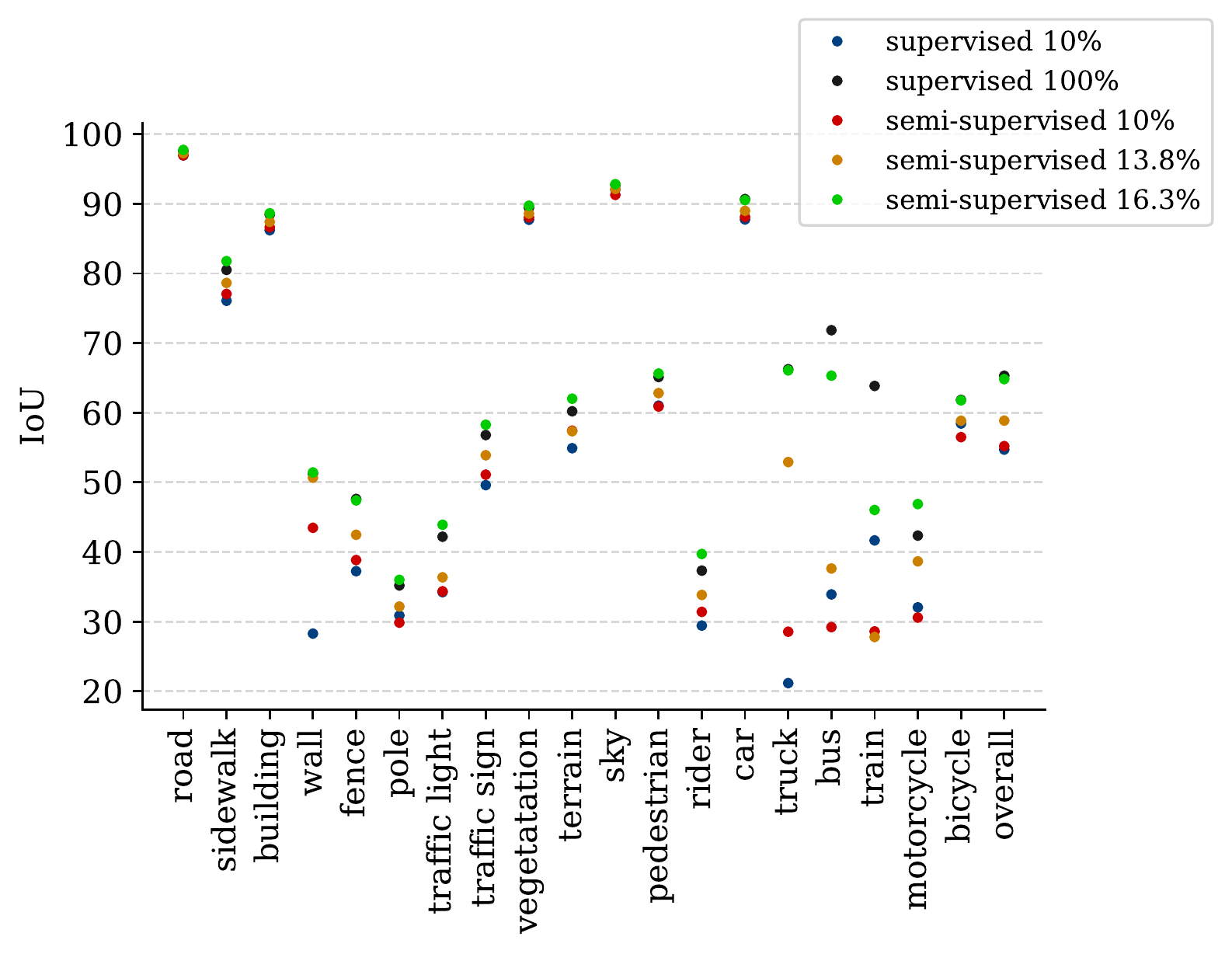}
    \end{center}
    \caption{CityScapes}
    \label{fig:results_cityscapes_main}
    \end{subfigure}
    \caption{\textbf{IoU:} Performance curves of using pseudo labels as a function of the total data used for training on (a) CamVid and (b) CityScapes. We reduce the labeling effort significantly, by requiring only $\sim$4\% and 6\% additional data on both datasets respectively. On the same initial data, the best state-of-the-art method achieves comparative performance with an additional 30\% data.}
\end{figure*}


\section{Experimental Setup and Discussions}
\subsection{Datasets}
\label{subsec:datasets}

We evaluate our proposed approach on CamVid and CityScapes datasets for semantic segmentation \cite{brostow2009semantic,cordts2016cityscapes}. We follow the widely adopted protocol for both datasets: we sample 10\% of the data as labeled data from the train set as our labeled data pool \cite{sinha2019variational,xie2020deal,ebrahimi2020minimax}. For simplicity and reproducibility, we sample images for the labeled set uniformly from the ground truth train set and consider all others from the train set as the unlabeled image pool. To ensure fairness in a comparison between all methods, we treat the ignore index for both datasets as part of the labeling information so that those areas are also potential regions to be acquisitioned for labeling.

\textbf{CityScapes} is a relatively larger dataset for semantic analysis of urban driving scenes at 1024 $\times$ 2048 resolution with 30 classes. It contains a total of 2975, 500, and 1525 images for training, validation and test respectively. We use the standard split of 2675, 300, and 500 images for training, validation, and test by replacing the validation set as test set and randomly sampling 300 images as validation. We downsample the images to 688 $\times$ 688 resolution and use 19 classes for training and evaluations, similar to  \cite{sinha2019variational,xie2020deal,ebrahimi2020minimax}.

\textbf{CamVid} is a driving scene understanding dataset consisting of images captured at 720 $\times$ 960 resolution with 32 classes. It contains a total of 701 images, with a split of 367, 101 and 233 images for training, validation, and test respectively. We use the widely adopted down sampled scenes at a resolution of 360 $\times$ 480 for our training and evaluation with 11 classes \cite{badrinarayanan2017segnet,equal2020,xie2020deal}.

\subsection{Experimental Configuration}
\label{subsec:expconfig}

We run our evaluations with MobileNetv2 as the backbone on the DeepLabv3+ semantic segmentation architecture \cite{Chen_2018_ECCV,Sandler_2018_CVPR}, a widely adopted standard for semantic segmentation in active learning. We adjust the MobileNetv2 backbone to have a higher stride of 16, similar to \cite{xie2020deal}. We also benchmark our results on the CityScapes dataset using DRNs \cite{Yu_2017_CVPR} to compare against \cite{sinha2019variational,ebrahimi2020minimax}, as they follow the same foundation approach in terms of initial labeled sets and image resolutions. We train all our networks with a batch size of four for 100 epochs and 200 epochs on the final stage of the active learning cycle. We start with an initial learning rate of 1$\times10^{-2}$ and use the ``poly"  learning schedule to gradually decrease the learning rate \cite{Chen_2018_ECCV,xie2020deal}. We set the value for the replay buffer size $M$ to 50 and 500 for CamVid and CityScapes, respectively. We sample four regions per image of 30 $\times$ 30 and 43 $\times$ 43 for two and five active learning cycles on CamVid and CityScapes, respectively. We refer to the supplemental for exact training details.

\textbf{Comparative Algorithms} We compare our methods against four other active selection methods as shown in Table \ref{tab:results_cityscapes}. These methods are designed for active learning on an image level basis - namely random selection, Entropy \cite{shannon2001mathematical}, Core-Set \cite{sener2018active}, and DEAL \cite{xie2020deal}. Attempts to compare with region level algorithms were partially unsuccessful because they use random sampling to initialize the labeled training set, which makes it challenging to replicate results. We refer the readers to the supplemental section for more details. 

\subsection{Main Results}
\label{subsec:mainresults}

On the CamVid dataset, we achieve 97\% of the performance (Fig. \ref{fig:results_camvid_main}) as compared to utilizing the full dataset with only 13.8\% of the labeled pixel data. The previous state-of-the-art approach achieved 94\% performance using 40\% of the data \cite{xie2020deal}. 

Our results on the CityScapes dataset are given in Table~\ref{tab:results_cityscapes}. Our approach outperforms existing methods on the CityScapes dataset using just 16\% of the labeled data as compared to 40\%. Except for the class Train, we improve the IoU scores on multiple tail classes (traffic light, traffic sign, truck, motorcycle and bicycle) by a significant amount. 

Figs. \ref{fig:results_camvid_main} and \ref{fig:results_cityscapes_main} show the gradual increase in IoU per class and overall as a function of data being utilized. Using semi-supervised learning, we improve the scores on both datasets by an initial 1-2 \% with the same amount of data. While this is lower than the gains seen in some other semi-supervised algorithms \cite{french2020semisupervised,Ouali_2020_CVPR,Chen_2021_CVPR,liu2021bootstrapping}, we believe that the limited number of parameters of MobileNetv2 act as a constraint as opposed to the much larger networks used in those studies. Irrespective of the initial boost, our approach iteratively increases the IoU on both datasets, while minimizing the overall labeling effort.

In addition to MobileNet, we also experiment with a comparatively lighter DRN-D-22 \footnote{We reached out to the corresponding paper's author for exact DRN version used in their experiments} network for comparisons with VAAL and MAL approaches for active learning \cite{Yu_2017_CVPR,sinha2019variational,ebrahimi2020minimax} and report our results in Table \ref{tab:comparisons_drn_approaches}. Direct comparisons are impossible due to the unknown labeled-unlabeled data split, and hence we report the results on a comparative basis with respect to the IoUs achieved with 10\%, maximum data sampled via active learning and fully supervised data pool. We make two key observations: 1) Training with semi-supervised learning on the initial 10\% data leads to an up-rise in the starting IoU, and 2) Our approach is able to achieve 97\% of the performance while using only 16\% of the data, as compared to 40\% and 30\% on VAAL and MAL respectively.

Finally, we compare our results to other region-based selection methods, namely EquAL and RALIS \cite{equal2020,Casanova2020Reinforced}. Again, direct comparison is not possible due to the unknown labeled-unlabeled split, so instead we compare performance using the same labeled fraction of the data. On CamVid, using 12\% of the data and a MobileNetv2 backbone, we achieved 95\% of the performance from the fully supervised training regime versus the 94\% that EquAL achieved using a relatively heavier ResNet-50 backbone with 12\% data and the 96\% that RALIS attained also using a ResNet-50 pretrianed on GTA and more (20\%) data. Our approach achieved an mIOU of 65.3 on CamVid, as compared to 63.4 from \cite{equal2020}, when starting with 8\% labeled data and with a budget of 12\% labeled data, using a ResNet-50 backbone with DeepLabv3+.

\subsection{Additional Results} We discuss the ablations with respect to our approach in this section. In all approach-specific experiments, we sample additional combinations of labeled-unlabeled combinations using random sampling over 5 runs and report our results as the mean and standard deviation over all experiments.

\begin{table}[t]
    \caption{\textbf{Discussions:} of additional studies conducted on CamVid and CityScapes dataset. We use MobileNetv2 for all our experiments.}
    \begin{subtable}[t]{0.45\textwidth}
        \def\arraystretch{1.2}  
        \small
        \centering
        \caption{\textbf{IoU:} with respect to different block sampling ratios on CamVid and CityScapes datasets.}
        \label{tab:regionsize_camvid_cityscapes}
        \resizebox{\textwidth}{!}{%
        \begin{tabular}{lllll}
        CamVid & mIoU & CityScapes & mIoU \\ \shline
        30 $\times$ 30 $\times$ 2 & 60.4 $\pm$ 1.4 & 43 $\times$ 43 $\times$ 2 & 61.8 $\pm$ 0.8\\
        30 $\times$ 30 $\times$ 4 & 61.4 $\pm$ 0.6 & 43 $\times$ 43 $\times$ 4 & 62.6 $\pm$ 2.2\\
        60 $\times$ 60 $\times$ 1 & 60.8 $\pm$ 2.4 & 86 $\times$ 86 $\times$ 1 & 61.4 $\pm$ 1.6 \\ \bottomrule
        \end{tabular}%
        }
     \end{subtable}
     \\
     \\
     \begin{subtable}[t]{0.3\textwidth}
        \def\arraystretch{1.2}  
        \small
        \centering
        \caption{\textbf{IoU:} with respect to different sampling schemes on CamVid and CityScapes datasets.}
        \label{tab:acquisition_fxs}
        \resizebox{\textwidth}{!}{%
        \begin{tabular}{@{}lll@{}}
         & \multicolumn{2}{c}{mIoU} \\ 
         & CamVid & CityScapes \\ \shline
        Random & 59.1 $\pm$ 1.8 & 59.8 $\pm$ 2.5 \\
        LS & 60.5 $\pm$ 0.5 & 60.3 $\pm$ 1.4  \\
        Ent & 61.2 $\pm$ 0.5 & 62.5 $\pm$ 1.8 \\
        Margin & 60.8 $\pm$ 0.6 & 61.8 $\pm$ 0.5 \\ \bottomrule
        \end{tabular}%
        }
     \end{subtable}
\end{table}

\textbf{Region vs Image:} Instead of using regions, we sample entire images based on the acquisition metric and follow with the semi-supervised learning approach where the newly acquired data is added to the labeled set of images. We repeat the active learning cycle 2 times, sampling 5\% of the queried image pool every time, resulting in a 20\% data usage. We observe that using regions benefit across both datasets (Tables \ref{tab:regionvsimage_camvid}, \ref{tab:regionvsimage_cityscapes}), often resulting in higher recall across tail distribution classes except for the `Train' class for CityScapes, which gets confused with `Bus'. We believe stronger regularizers can help prevent pseudo label confusion and mitigate this performance gap.

\textbf{Confidence Weighting:} benefits both datasets, most noticeable in the very first active learning cycle (CamVid: $57.2 \longrightarrow 58.5$, CityScapes: $56.15 \longrightarrow 56.6$). This is important as having the correct knowledge and not falling to the dataset bias is crucial towards obtaining the perfect set of samples. More specifically, for CityScapes, we further observe an average difference of nearly 5\% for tail classes - traffic light, traffic sign, pedestrian, rider, and motorcycle - which implies a lot of potential false label associations were averted by imposing a simple constraint on the pseudo labels.

\textbf{Balanced Classmix:} also benefits both datasets, and this is most noticable in the final active learning cycle (CamVid: $60.4 \longrightarrow 61.5$, CityScapes: $61.3 \longrightarrow 63.15$). Active learning performs significantly better with using Balanced Classmix - depending on the initial data pool, we observe more inconsistencies in the final results if Balanced Classmix is not used as it compensates for lack of enough contributions from the tail classes. This also comes as a little deterrent, specially for the `Train' category, as this same regularizer works negatively to over-sample areas that confuse the category.

\textbf{Region Size:} We experiment with different region sizes for both datasets in Table \ref{tab:regionsize_camvid_cityscapes}. Statistically, there are multiple combinations possible, and we achieve the best consistent results across our final size of choosing.

\textbf{Sampling Strategy:} We conclude our study of hyperparameters in the experiment with the metric used for sampling the regions within an image for labeling (Table \ref{tab:acquisition_fxs}). The values indicate that softmax entropy works the best for our approach, with softmax margin performing nearly well. This can also be explained by the Confidence Weighting that enforces a tougher constraint and thus makes identifying key areas possible.

\section{Conclusions}
\label{sec:conclusion}

We visit active learning for semantic segmentation with the goal of incorporating semi-supervision based pseudo labeling into each training cycle to reduce labeling costs. We propose two regularizers to prevent issues with dataset bias in terms of head and tail classes, Confidence Weighting and Balanced Classmix. Our approach achieves comparable performance to full supervised data with a significant reduction in the amount of labels used. A current limitation of our work is lack of understanding with respect to classes that have not been seen during training (open-set or heavily under-represented), however, we believe that our work can set a precedent for further research in the area.

\section*{Acknowledgements} This work was supported by the Dynamic Data Driven Applications Systems Program, Air Force Office of Scientific Research under Grant FA9550-19-1-0021. We gratefully acknowledge the support of NVIDIA Corporation with the donations of the Titan X and Titan Xp Pascal GPUs used for this research, and DataCrunch.io for the usage of Nvidia V100 GPUs.


{\small
\bibliographystyle{ieee_fullname}
\bibliography{dlcitations}
}

\clearpage
\begin{center}
    {\Large Supplemental Material \normalsize}
\end{center}
\beginsupplement

We will release all code towards reproducing the results in this paper post publication. We discuss the experiments and related results in the following sections.

\section{Experimental Configurations}
\begin{table}
\small
\caption{Supervised Learning parameters.}
\label{tab:supervised_settings}
\def\arraystretch{1.2}
\centering
\resizebox{0.45\textwidth}{!}{%
\begin{tabular}{@{}lll@{}}
Config & CamVid & CityScapes \\ \shline
Learing Rate & $1e-2$ & $1e-2$ \\
Optimizer & SGD & SGD \\
Scheduler & PolyLR & PolyLR \\
 &  &  \\
Batch Size & 4 & 4 \\
Epoch Iterations & 100 & 200 \\
Epochs & 100 & 100 \\
 &  &  \\
Augmentations & \begin{tabular}[c]{@{}l@{}}Resize({[}0.75,1.25{]}),\\ ColorJitter(p = 0.5),\\ HFlip(p = 0.5)\end{tabular} & \begin{tabular}[c]{@{}l@{}}Resize({[}0.5,2.0{]}),\\ ColorJitter(p = 0.5),\\ HFlip(p = 0.5)\end{tabular} \\ \bottomrule
\end{tabular}%
}
\end{table}

\begin{table}[t]
\small
\caption{Active Learning with Semi-Supervised Learning parameters.}
\label{tab:semisup_al_settings}
\def\arraystretch{1.2}
\centering
\resizebox{0.45\textwidth}{!}{%
\begin{tabular}{@{}lll@{}}
Config & CamVid & CityScapes \\ \shline
Learing Rate & 1e-2 & 1e-2 \\
Optimizer & SGD & SGD \\
Scheduler & PolyLR & PolyLR \\
 &  &  \\
Batch Size (Labeled) & 4 & 4 \\
Batch Size (Un-Labeled) & 4 & 4 \\
Epoch Iterations & 50 & 100 \\
Epochs & 100 & 100 \\
Coldstart Epochs & 10 & 10 \\
Final Cycle Epochs & 200 & 200 \\
 &  &  \\
Active Learning Cycles & 2 & 5 \\
Active Learning Regions & 30 $\times$ 30 $\times$ 4 & 43 $\times$ 43 $\times$ 4 \\
 &  &  \\
\begin{tabular}[c]{@{}l@{}}Augmentations\\ (Labeled)\end{tabular} & \begin{tabular}[c]{@{}l@{}}Resize({[}0.75,1.25{]}),\\ ColorJitter(p = 0.5),\\ HFlip(p = 0.5)\end{tabular} & \begin{tabular}[c]{@{}l@{}}Resize({[}0.5,2.0{]}),\\ ColorJitter(p = 0.5),\\ HFlip(p = 0.5)\end{tabular} \\
 &  &  \\
\begin{tabular}[c]{@{}l@{}}Weak Augmentations\\ (Un-Labeled)\end{tabular} & \begin{tabular}[c]{@{}l@{}}Resize({[}0.75,1.25{]}),\\ HFlip(p = 0.5)\end{tabular} & \begin{tabular}[c]{@{}l@{}}Resize({[}0.5,2.0{]}),\\ HFlip(p = 0.5)\end{tabular} \\
 &  &  \\
\begin{tabular}[c]{@{}l@{}}Strong Augmentations\\ (Un-Labeled)\end{tabular} & \begin{tabular}[c]{@{}l@{}}Resize({[}0.75,1.25{]}),\\ ColorJitter(p = 0.5),\\ HFlip(p = 0.5),\\ ClassMix\cite{Olsson_2021_WACV}\end{tabular} & \begin{tabular}[c]{@{}l@{}}Resize({[}0.5,2.0{]}),\\ ColorJitter(p = 0.5),\\ HFlip(p = 0.5),\\ ClassMix\cite{Olsson_2021_WACV}\end{tabular} \\
 &  &  \\
Replay Buffer Size & 50 & 500 \\ \bottomrule
\end{tabular}%
}

\end{table}

Tables \ref{tab:supervised_settings}, \ref{tab:semisup_al_settings} show the default hyper-parameters for training supervised and active learning based networks. We use the standard set of augmentations followed for supervised and semi-supervised learning, and append the latter with ClassMix \cite{Olsson_2021_WACV}. For semi-supervised learning per active learning cycle, we train with only labeled data for the initial 10 epochs to provide a good initialization for the teacher. In addition, we only start adaptive ClassMix once the first active learning cycle has completed - this reduces the initial training time and also makes the intended usage with additional labeled data. We train our networks on two Nvidia Titan Xps (CamVid) and two Nvidia V100s (CityScapes).

\section{Comparison to Region-based approaches}

We consider two algorithms developed for region-based semantic segmentation - RALIS \cite{Casanova2020Reinforced} and EquAL \cite{equal2020}. RALIS uses a ResNet-50 backbone with Feature Pyramid Networks \cite{lin2017feature}, enhanced with pretraining on the GTA-V dataset \cite{richter2016playing}. It also uses initial labeled sets of 30\% for CamVid and 12\% for CityScapes, which are higher than our initial budget of 10\% on both datasets. In comparison, our approach achieved 97\% of its fully-supervised performance on CamVid using while only actively sampling an additional 3.8\% of total data (13.8\% overall), whereas RALIS required an additional 24\% of the total data (54\% overall) to reach a maximum performance of 96\%. In addition, our approach achieved nearly identical results to the fully-supervised performance on the CityScapes dataset using only an additional 8\% of the total data (18\% overall), whereas RALIS achieved 96\% of fully-supervised performance using an additional 9\% of total data (21\% overall). It is worth mentioning that in both datasets of interest, RALIS used a pretraining boost with the GTA V dataset on ResNet-50, whereas we only use ImageNet-pretrained weights on MobileNetv2.

For EquAL, direct comparison is not possible due to the unknown labeled-unlabeled split, so instead we compare performance using the same labeled fraction of the data with ResNet-50 backbone. Under the same training paradigm (starting with 8\% labeled data, a budget of 12\% labeled data, and using a ResNet-50 backbone with DeepLabv3+), our approach achieved an mIOU of 65.3 $\pm$ 0.2 on CamVid, as compared to 63.4 from \cite{equal2020} \footnote{as mentioned in EquAL's GitHub repository}. For CityScapes, we found it realistically impossible to begin with only 1\% labeled data due to sampling concerns and no recorded data splits, so we begin with 3.5\% data instead which is a conventional choice for semi-supervised learning tasks \cite{alonso2021semi,Lai_2021_CVPR,Chen_2021_CVPR,liu2021bootstrapping}. Compared to EquAL's 12\% usage with an mIoU of 67.4, we obtained 66.7 $\pm$ 1.5 using only 10\% of the total labeled data. We believe the higher variance here as opposed to our other results is caused by using only 3.5\% data initially labeled data (vs. 10\% in the other experiments), and further research could help reduce this variance with improvements in SSL \cite{alonso2021semi,Lai_2021_CVPR,Chen_2021_CVPR,liu2021bootstrapping}.

\section{Visual Results}
\begin{figure*}[t]
    \centering
    \begin{subfigure}{0.35\textwidth}
        \begin{center}
        \includegraphics[width=\linewidth]{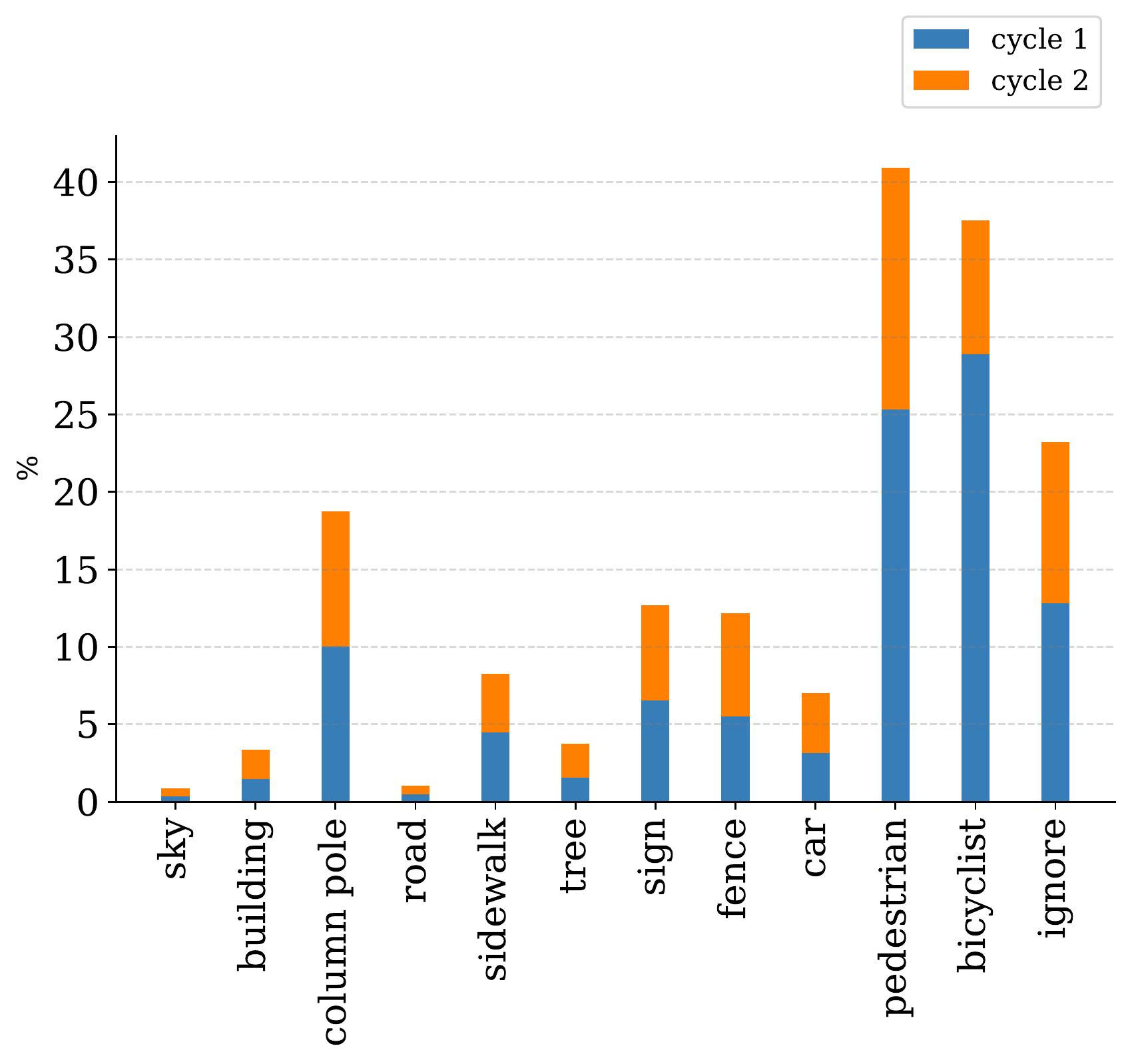}
        \end{center}
        \caption{CamVid}
        \label{fig:selection_dataset_camvid}
    \end{subfigure}
    \hfill
    \begin{subfigure}{0.55\textwidth}
        \begin{center}
        \includegraphics[width=\linewidth]{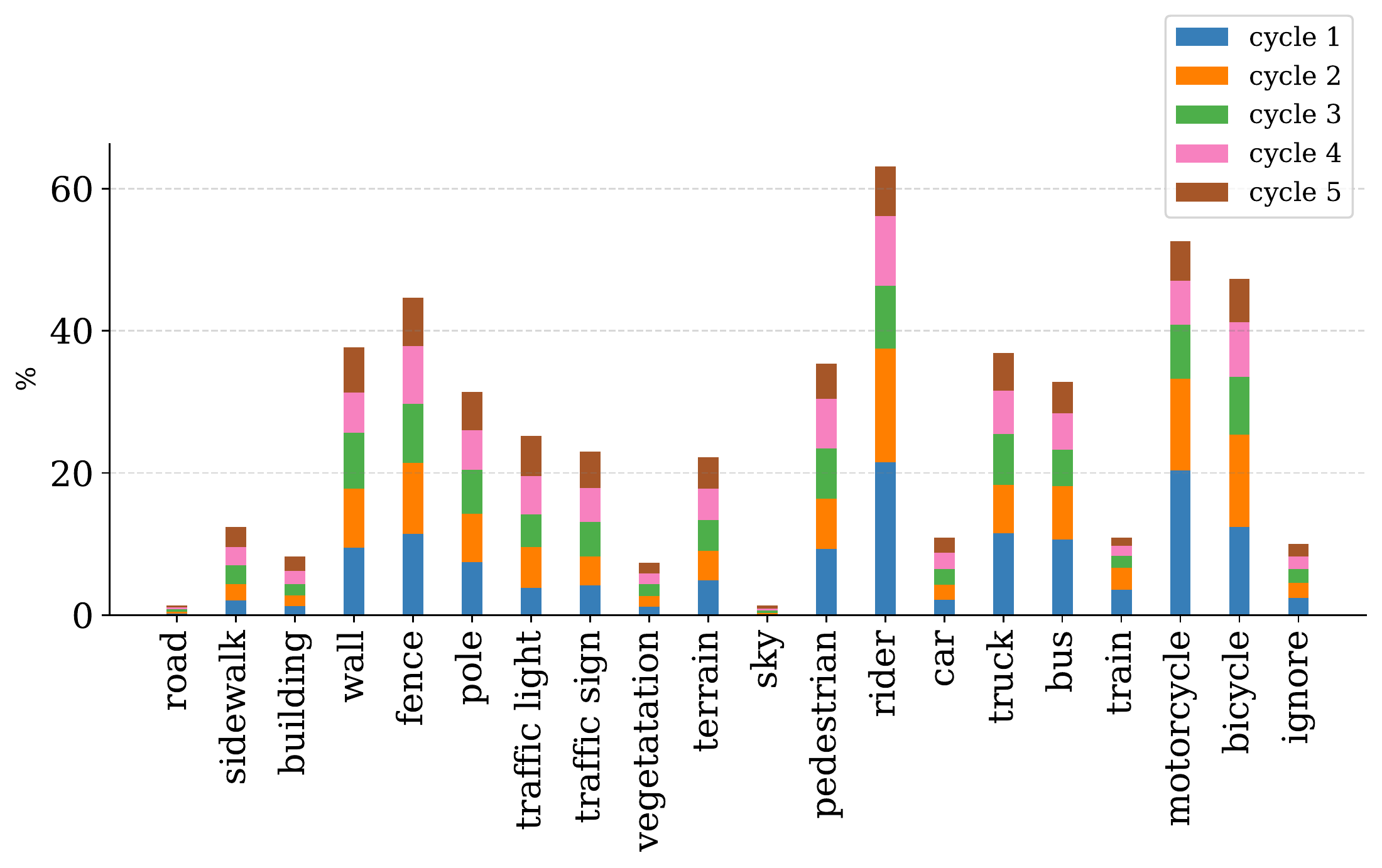}
        \end{center}
        \caption{CityScapes}
        \label{fig:selection_dataset_cityscapes}
    \end{subfigure}
\caption{What does the network want? We visualize the region-wise samples per active learning cycle on both datasets for our main split}
\label{fig:selection_dataset}
\end{figure*}
\begin{figure*}[t]
    \centering
    \begin{subfigure}[t]{0.24\textwidth}
        \begin{center}
        \includegraphics[width=\linewidth]{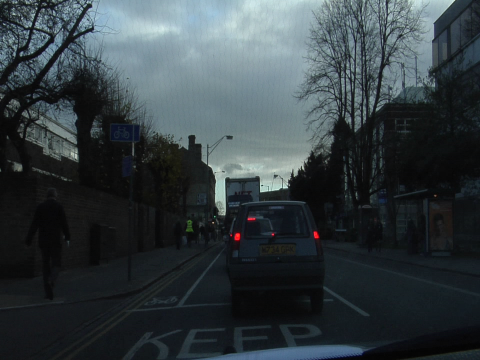}
        \end{center}
    \end{subfigure}
    \hfill
    \begin{subfigure}[t]{0.24\textwidth}
        \begin{center}
        \includegraphics[width=\linewidth]{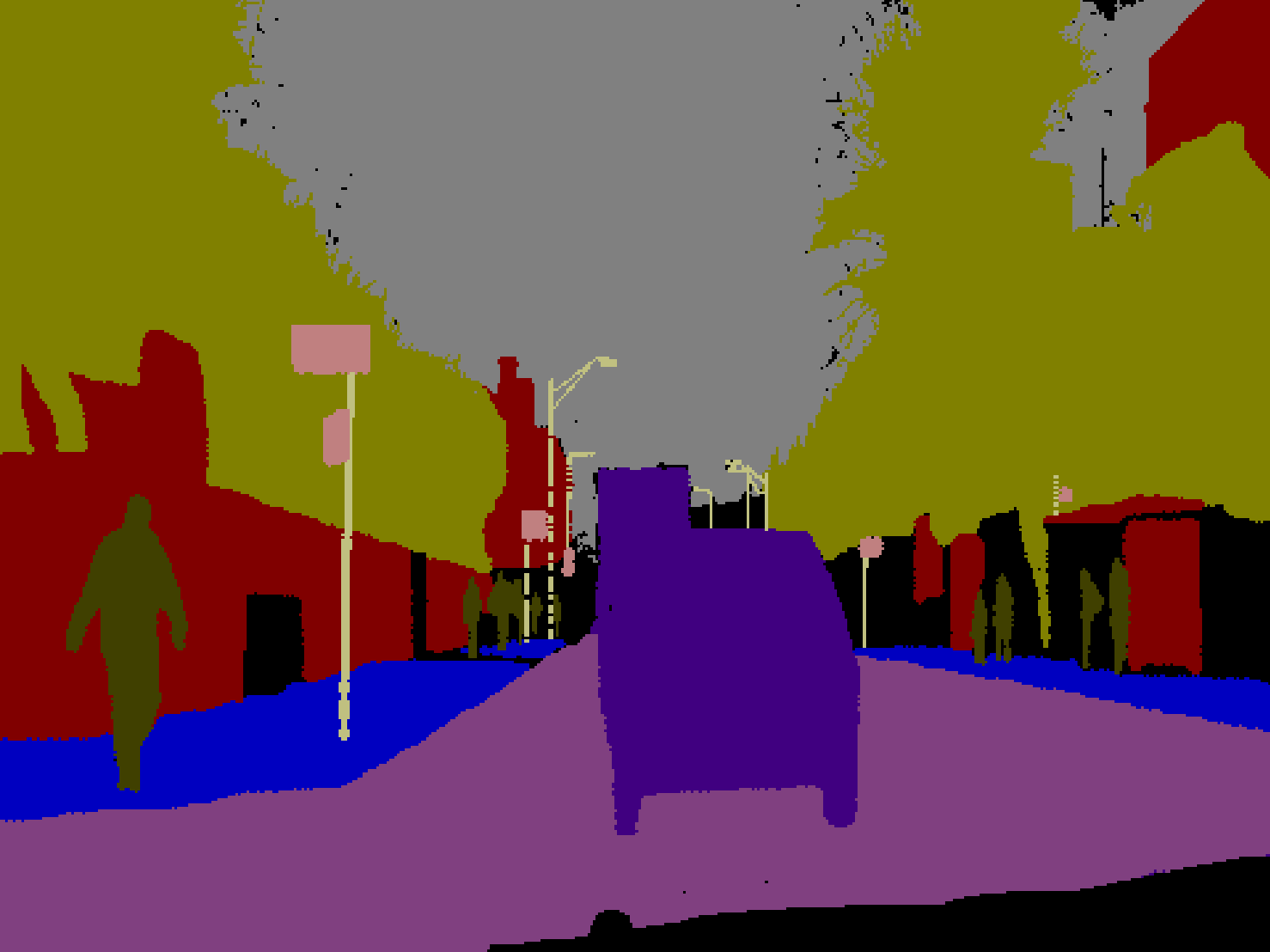}
        \end{center}
    \end{subfigure}
    \hfill
    \begin{subfigure}[t]{0.24\textwidth}
        \begin{center}
        \includegraphics[width=\linewidth]{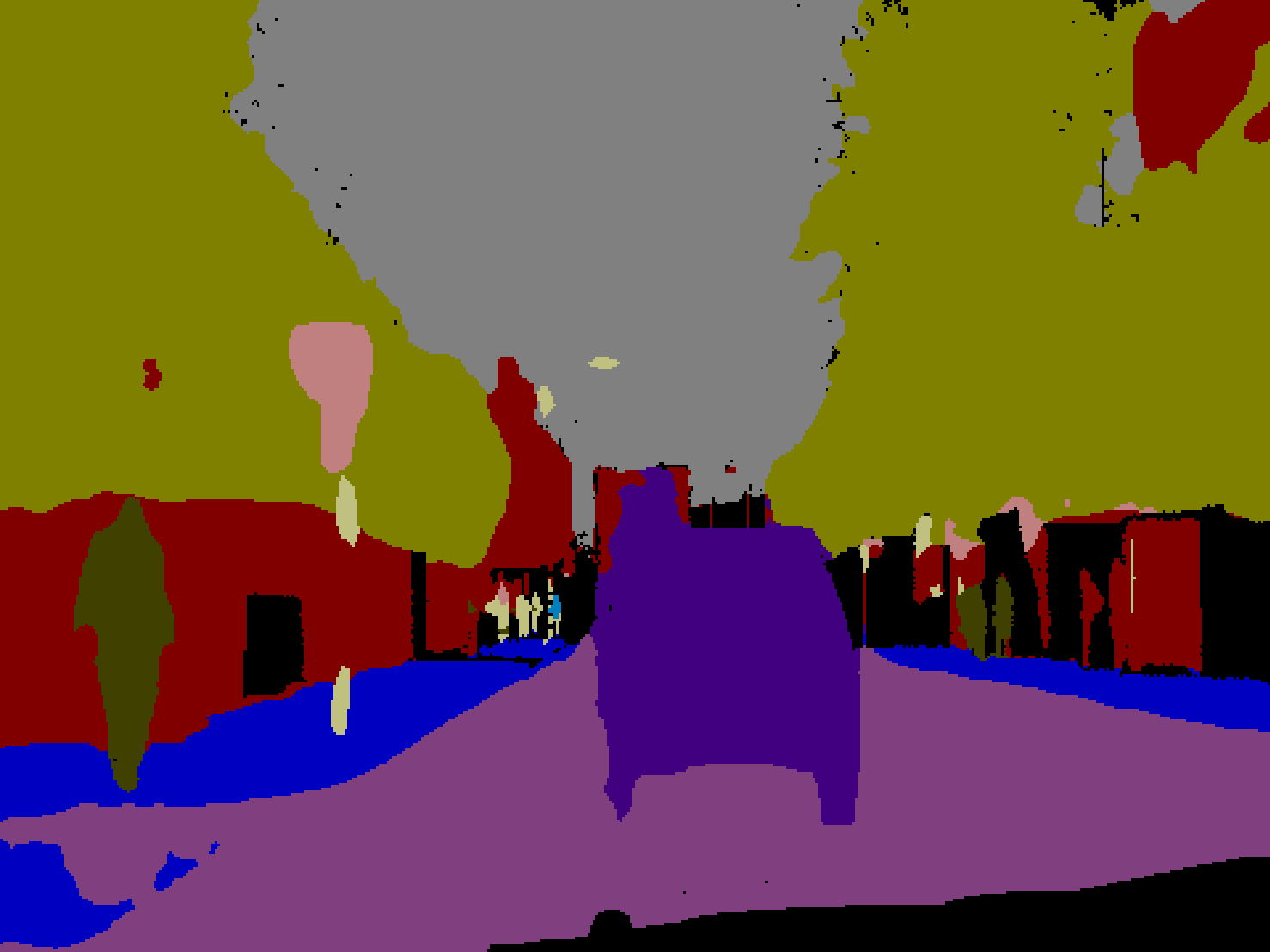}
        \end{center}
    \end{subfigure}
    \hfill
    \begin{subfigure}[t]{0.24\textwidth}
        \begin{center}
        \includegraphics[width=\linewidth]{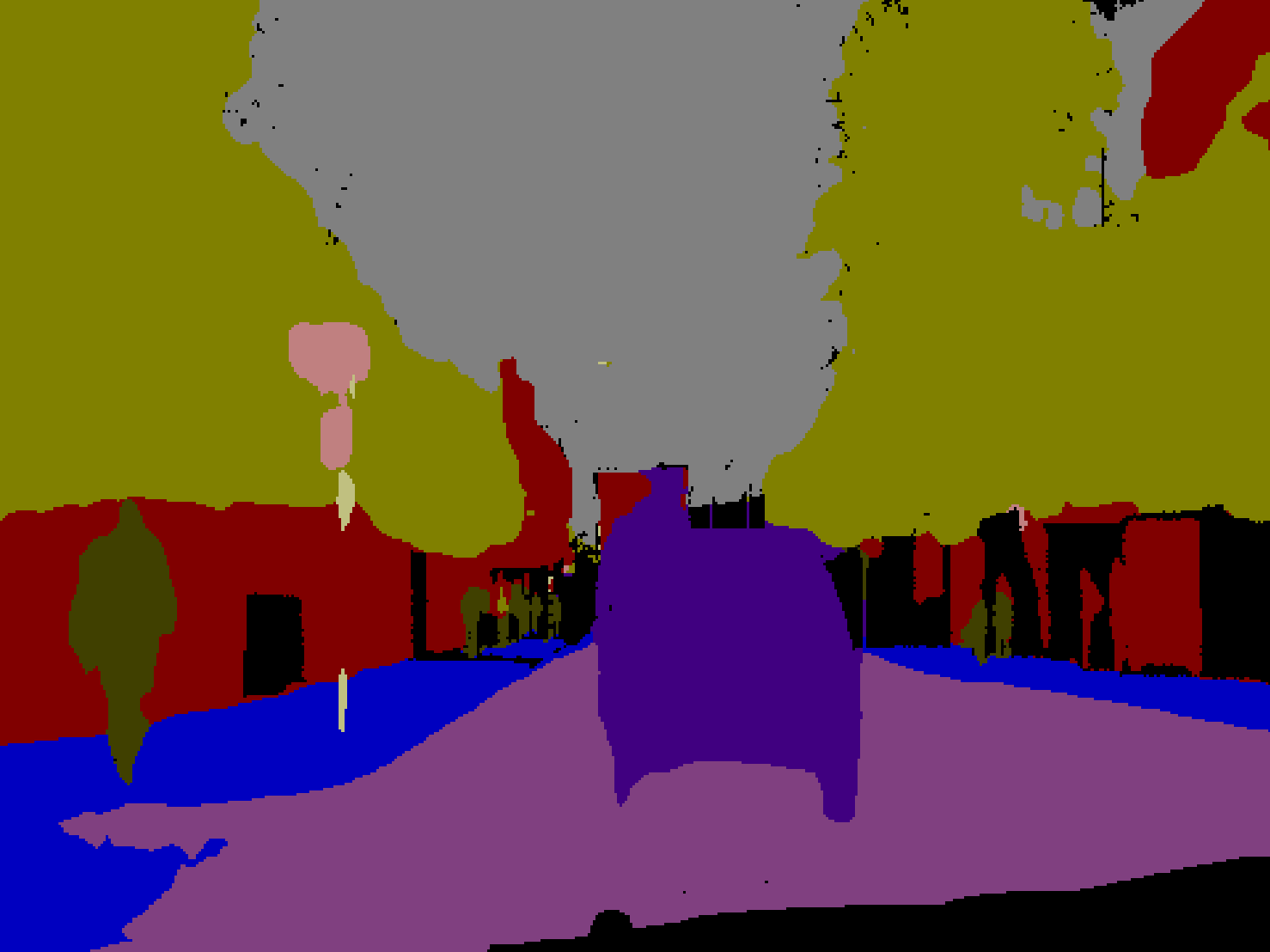}
        \end{center}
    \end{subfigure}
    \vspace*{\fill}
    \begin{subfigure}[t]{0.24\textwidth}
        \begin{center}
        \includegraphics[width=\linewidth]{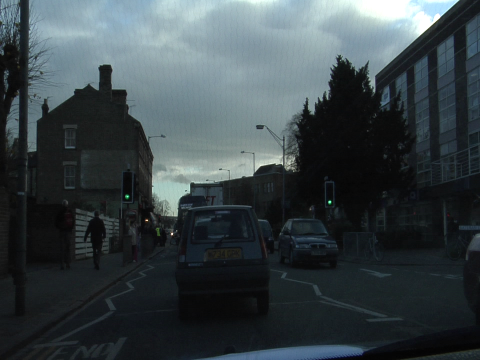}
        \end{center}
    \end{subfigure}
    \hfill
    \begin{subfigure}[t]{0.24\textwidth}
        \begin{center}
        \includegraphics[width=\linewidth]{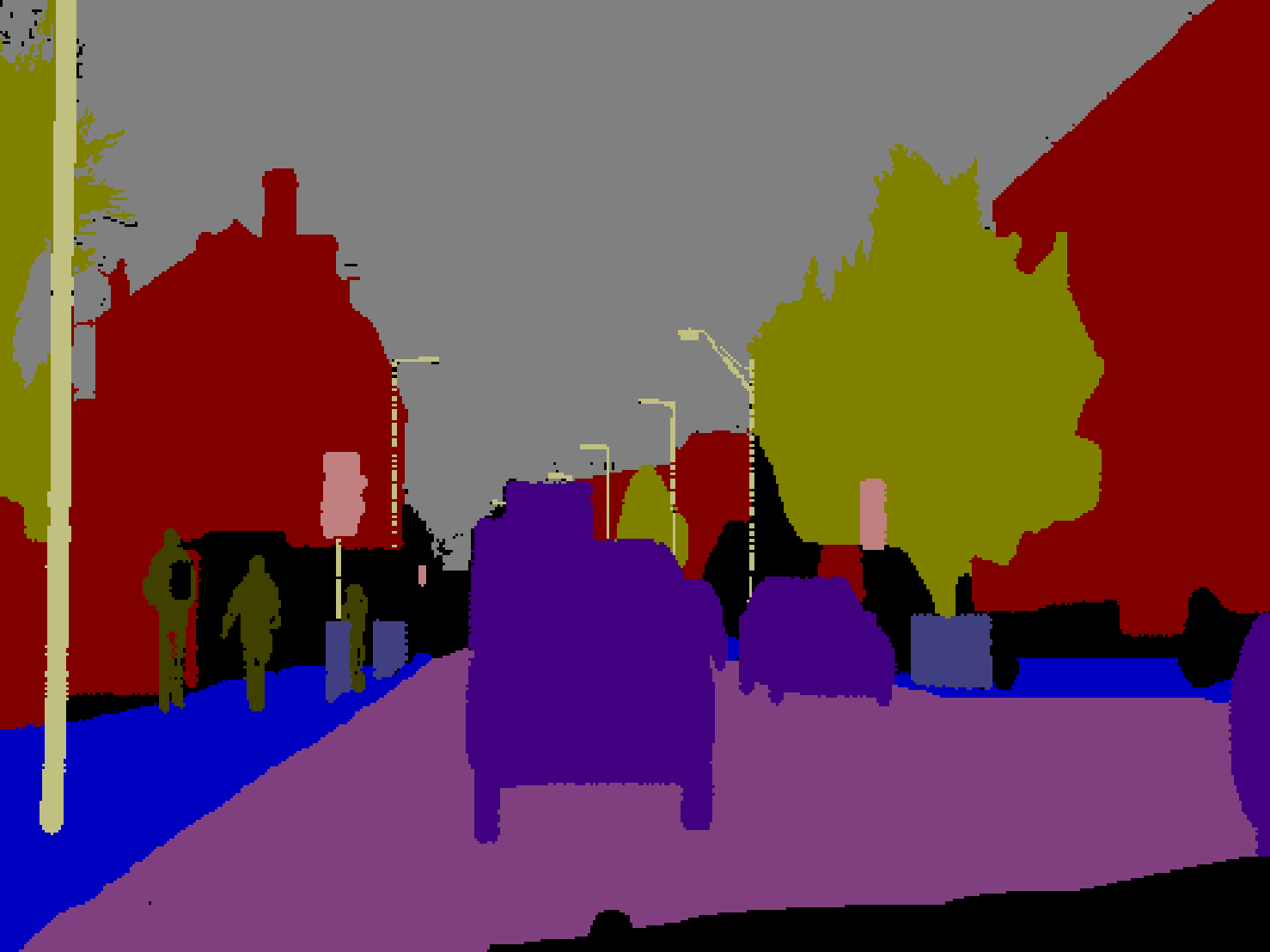}
        \end{center}
    \end{subfigure}
    \hfill
    \begin{subfigure}[t]{0.24\textwidth}
        \begin{center}
        \includegraphics[width=\linewidth]{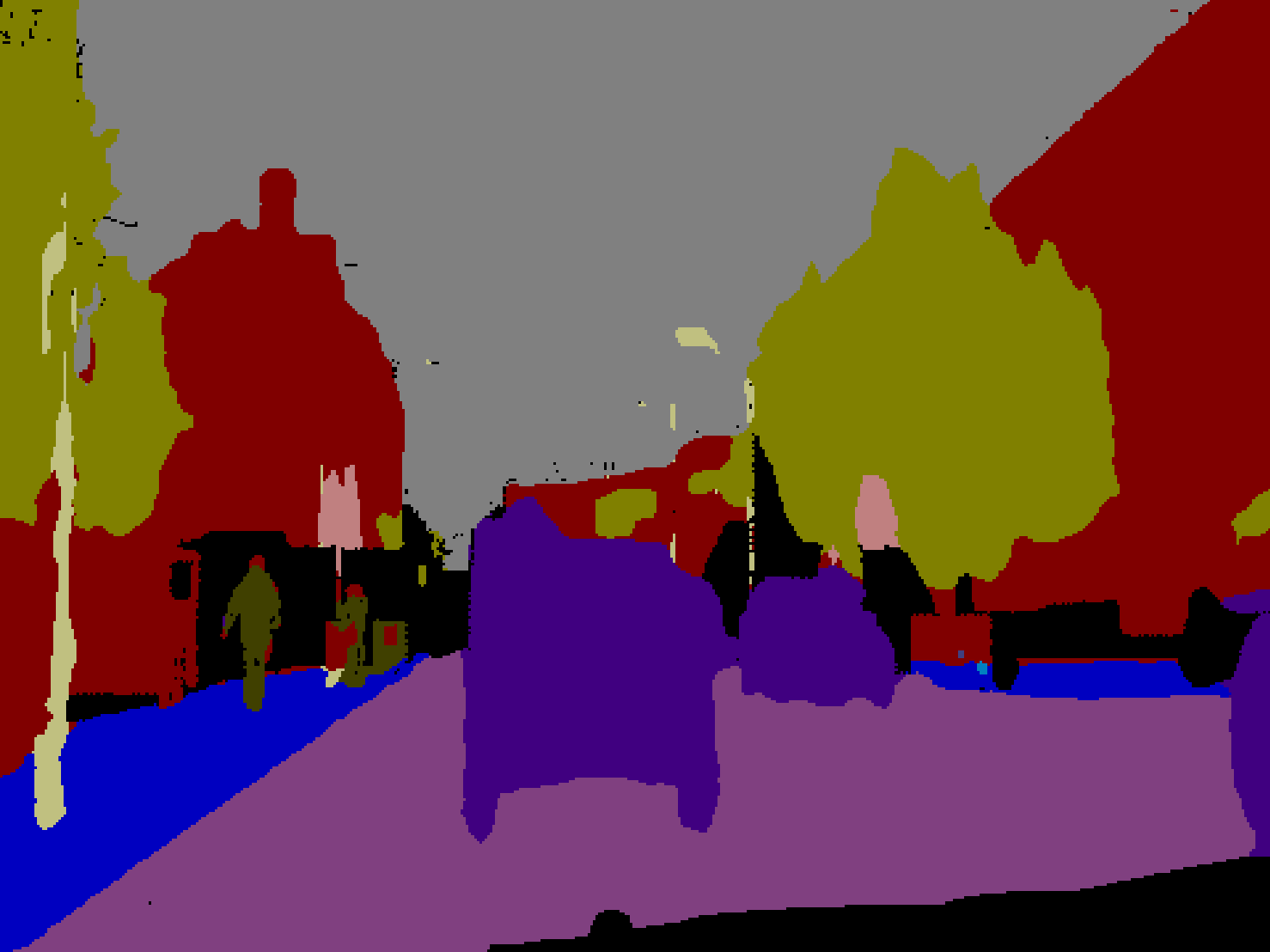}
        \end{center}
    \end{subfigure}
    \hfill
    \begin{subfigure}[t]{0.24\textwidth}
        \begin{center}
        \includegraphics[width=\linewidth]{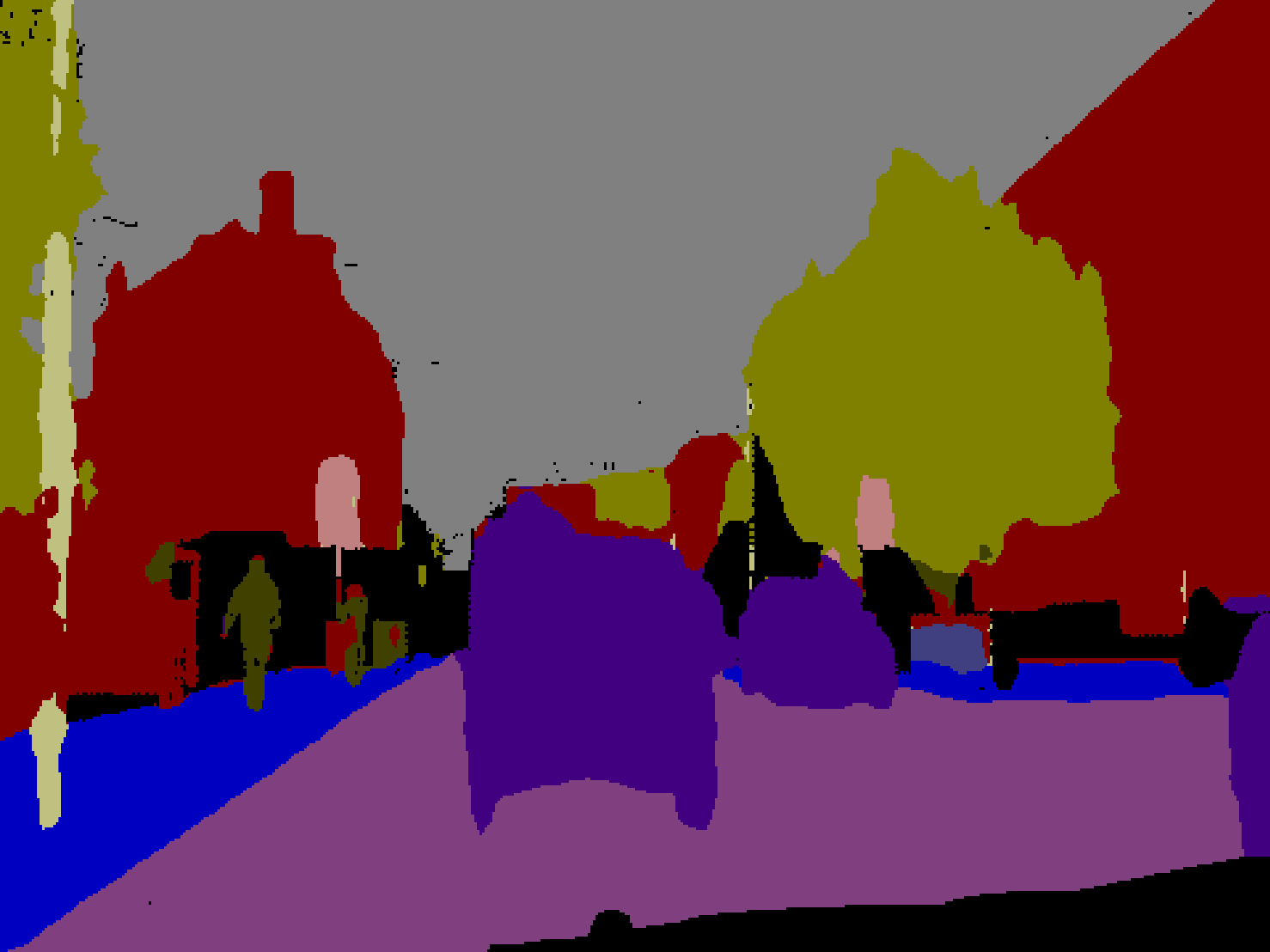}
        \end{center}
    \end{subfigure}
    \vspace*{\fill}
    \begin{subfigure}[t]{0.24\textwidth}
        \begin{center}
        \includegraphics[width=\linewidth]{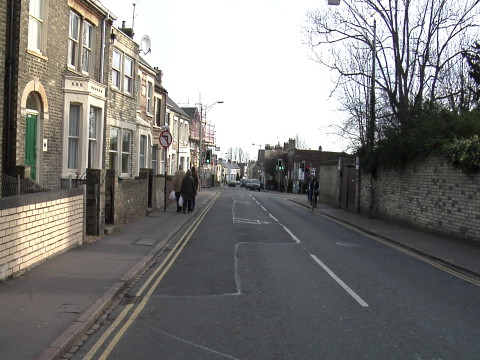}
        \end{center}
    \end{subfigure}
    \hfill
    \begin{subfigure}[t]{0.24\textwidth}
        \begin{center}
        \includegraphics[width=\linewidth]{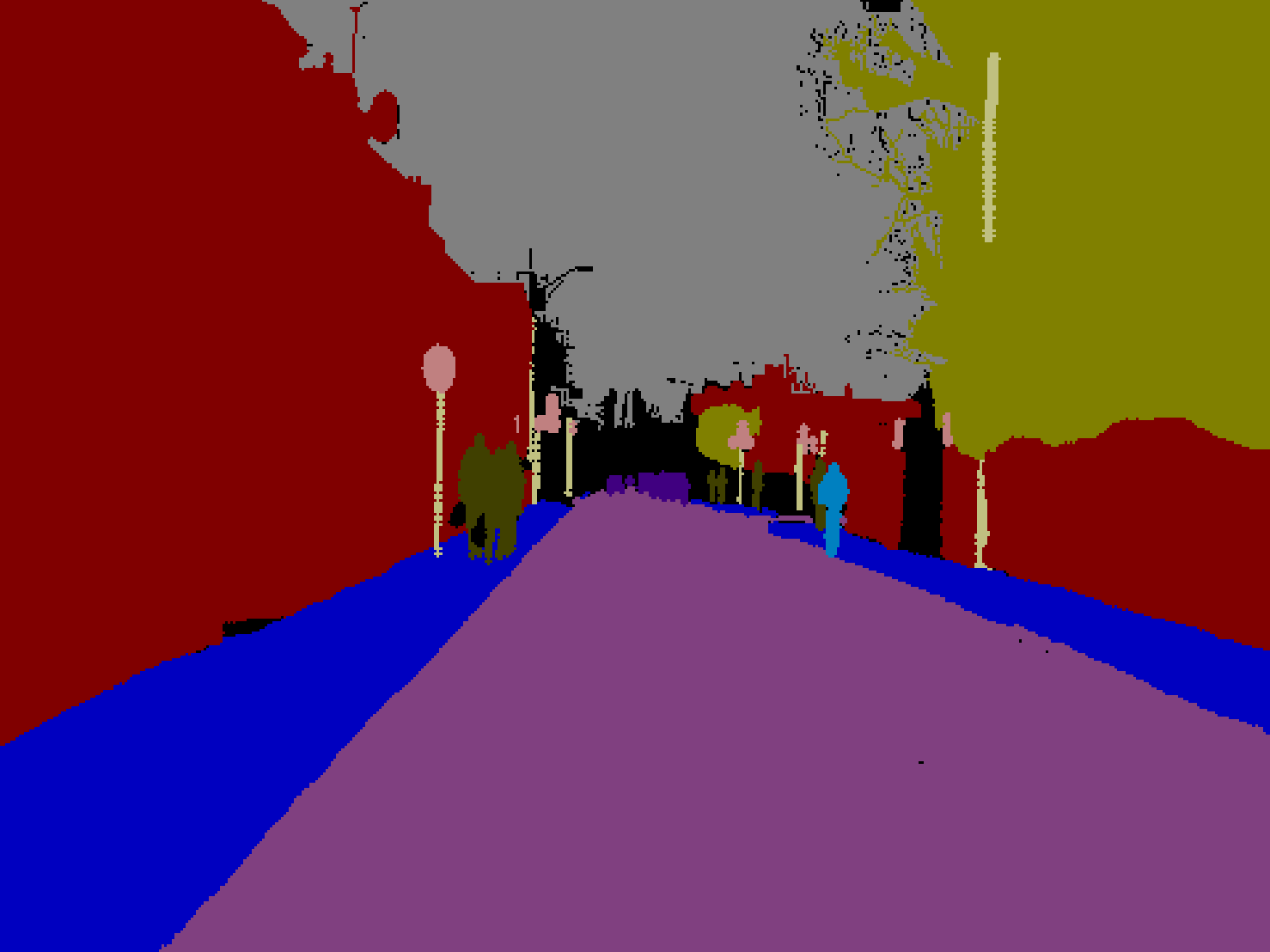}
        \end{center}
    \end{subfigure}
    \hfill
    \begin{subfigure}[t]{0.24\textwidth}
        \begin{center}
        \includegraphics[width=\linewidth]{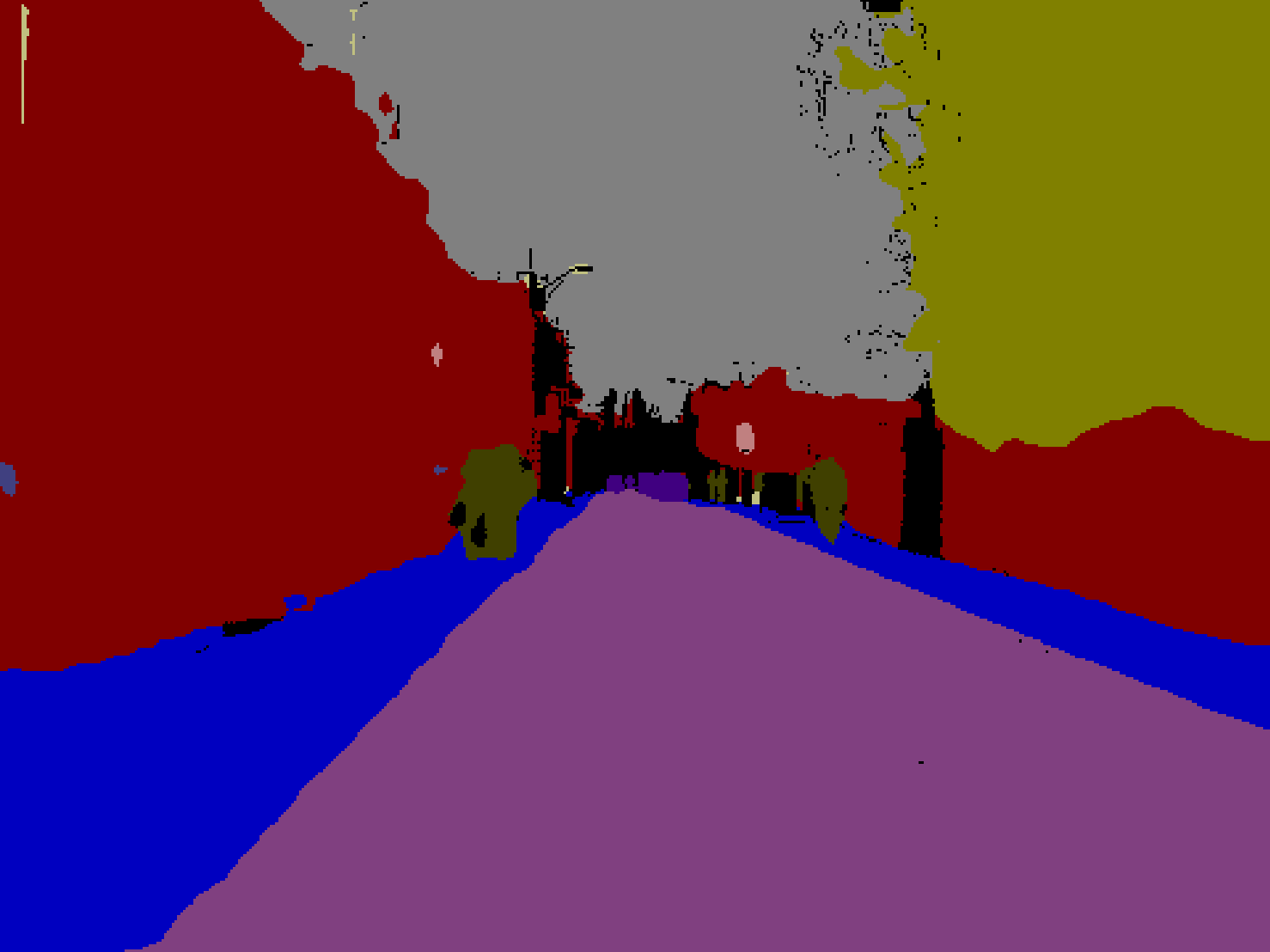}
        \end{center}
    \end{subfigure}
    \hfill
    \begin{subfigure}[t]{0.24\textwidth}
        \begin{center}
        \includegraphics[width=\linewidth]{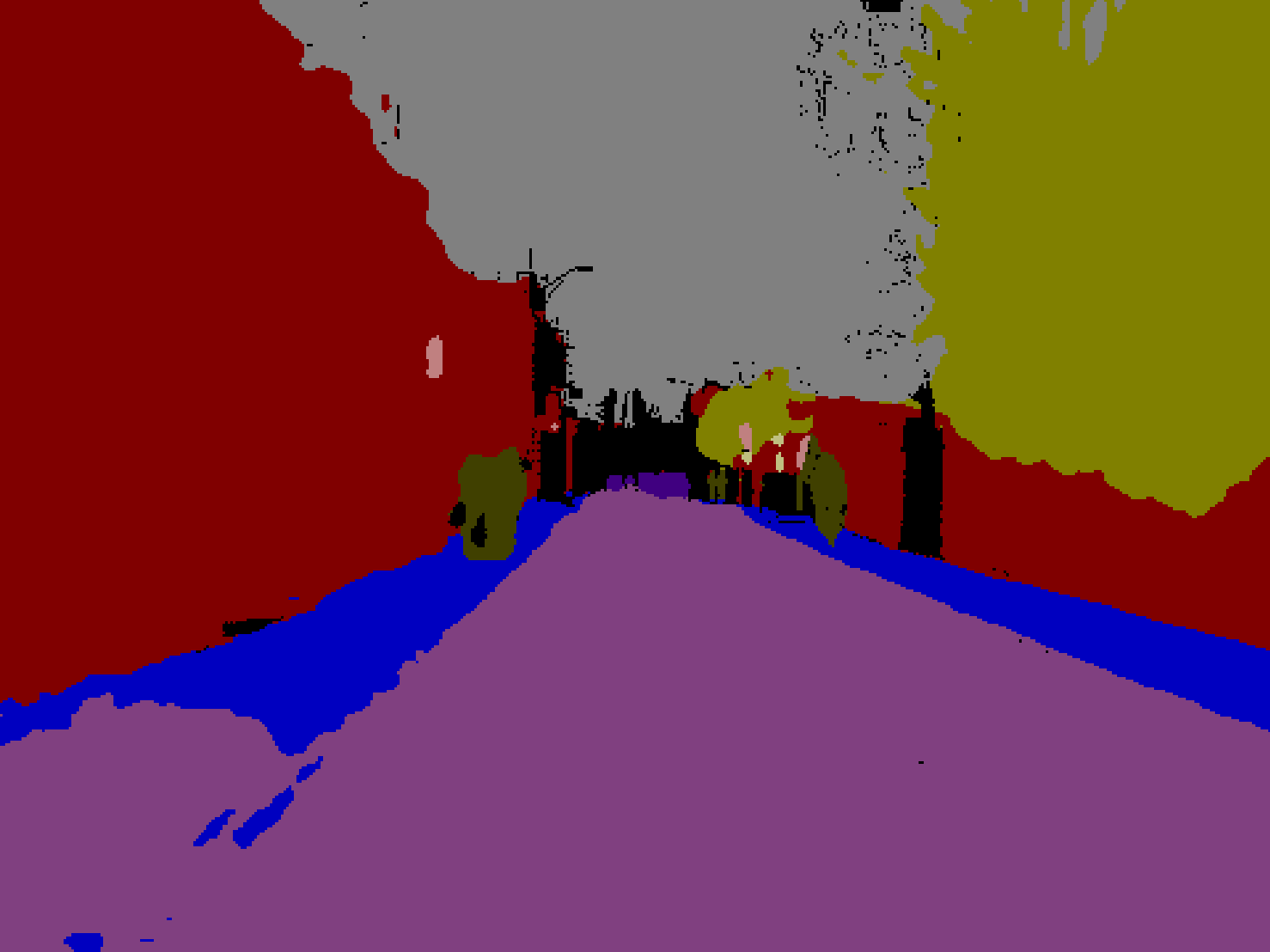}
        \end{center}
    \end{subfigure}
    \vspace*{\fill}
    \begin{subfigure}[t]{0.24\textwidth}
        \begin{center}
        \includegraphics[width=\linewidth]{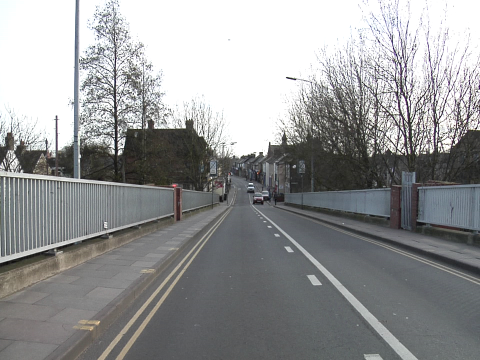}
        \end{center}
    \end{subfigure}
    \hfill
    \begin{subfigure}[t]{0.24\textwidth}
        \begin{center}
        \includegraphics[width=\linewidth]{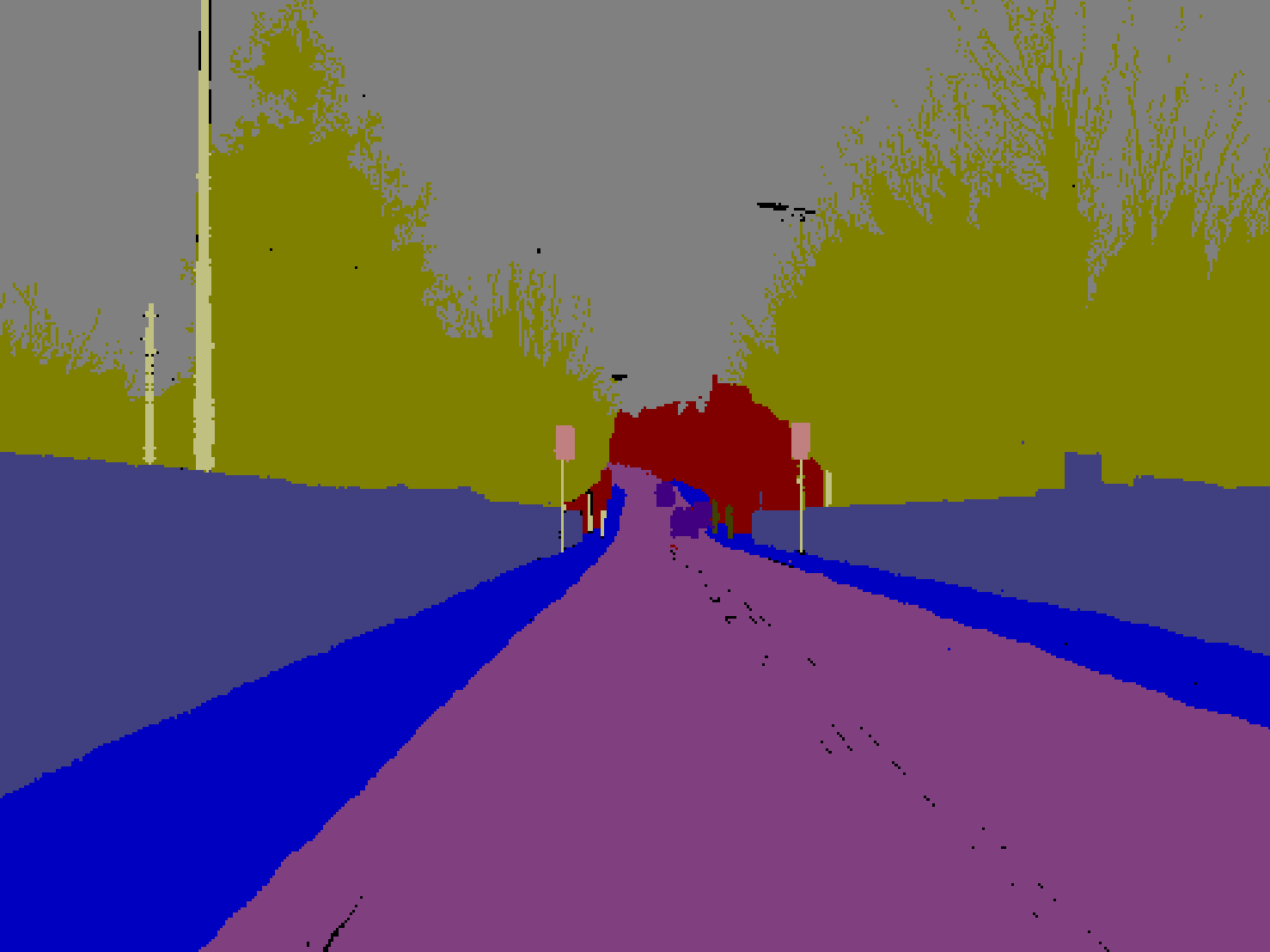}
        \end{center}
    \end{subfigure}
    \hfill
    \begin{subfigure}[t]{0.24\textwidth}
        \begin{center}
        \includegraphics[width=\linewidth]{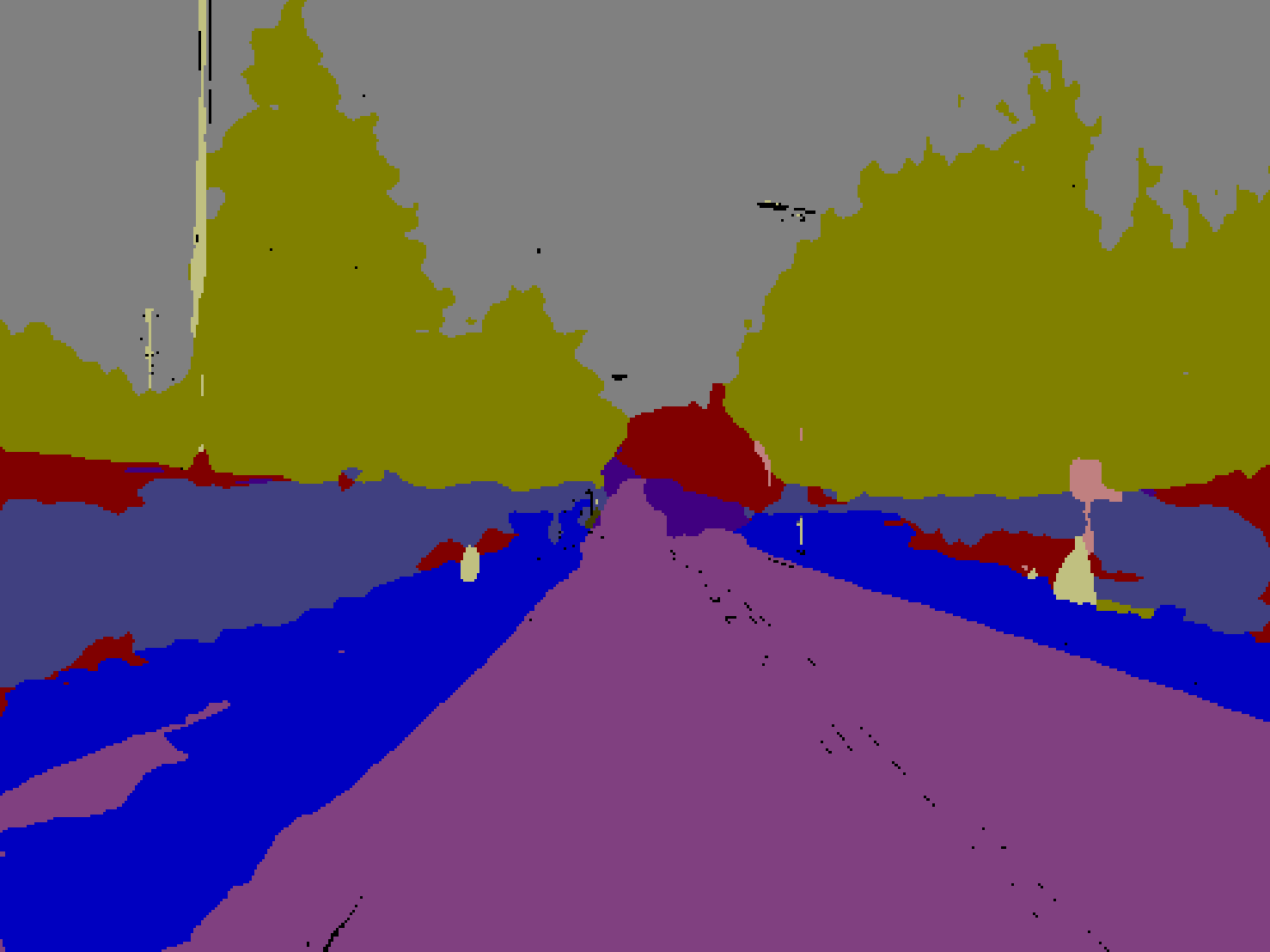}
        \end{center}
    \end{subfigure}
    \hfill
    \begin{subfigure}[t]{0.24\textwidth}
        \begin{center}
        \includegraphics[width=\linewidth]{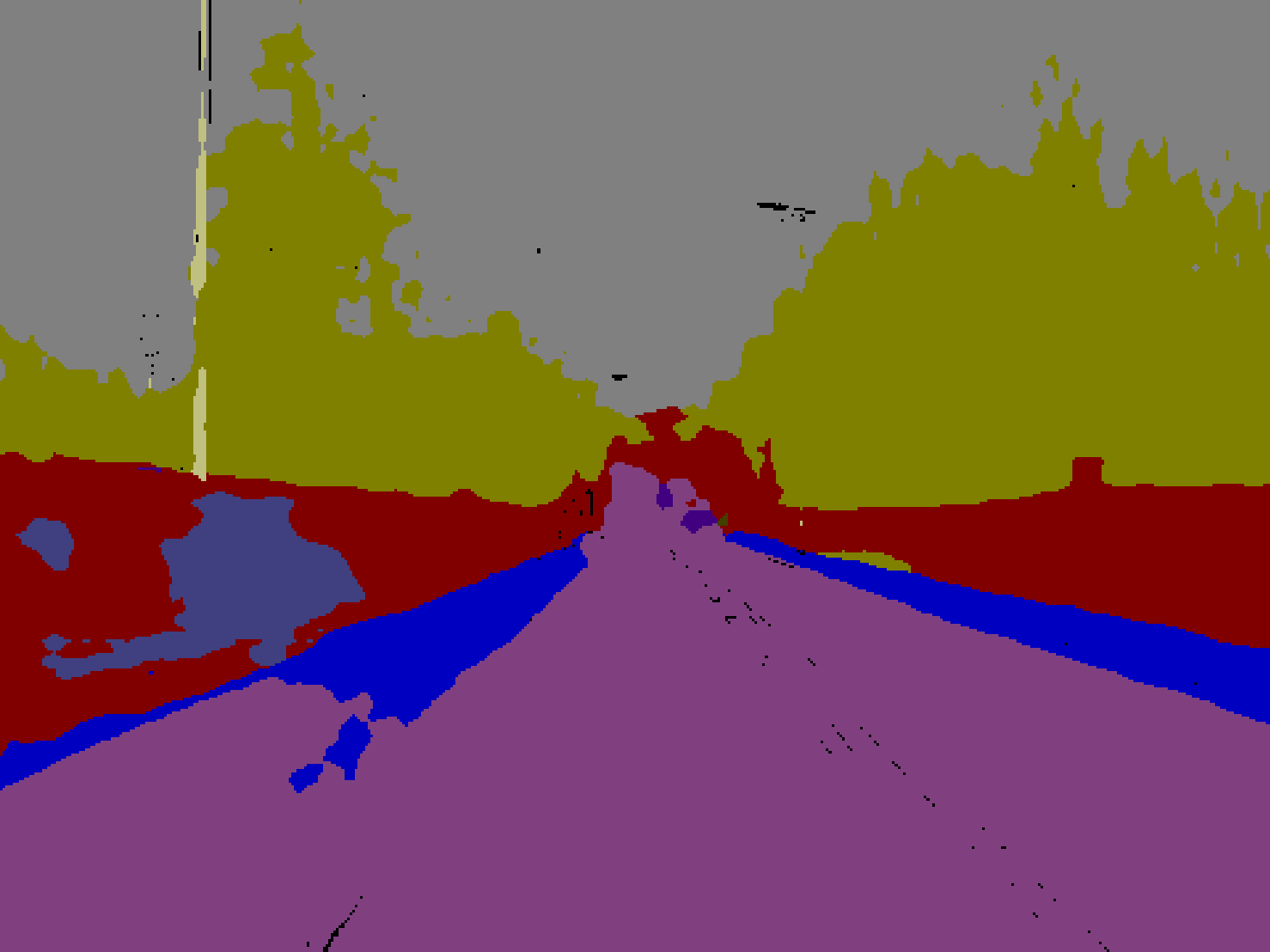}
        \end{center}
    \end{subfigure}
\caption{Examples of semantic segmentation outputs on CamVid, the columns represent the image (left), the ground truth (center-left), the predictions of a supervised network (center-right), and the predictions of a network trained via S4AL (right).}
\label{fig:viz_dataset_camvid}
\end{figure*}

\begin{figure*}[t]
    \centering
    \begin{subfigure}[t]{0.24\textwidth}
        \begin{center}
        \includegraphics[width=\linewidth]{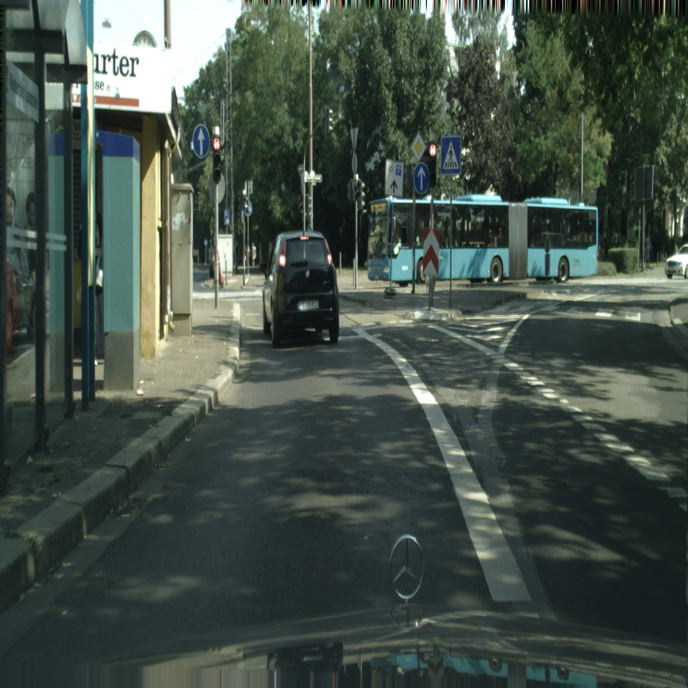}
        \end{center}
    \end{subfigure}
    \hfill
    \begin{subfigure}[t]{0.24\textwidth}
        \begin{center}
        \includegraphics[width=\linewidth]{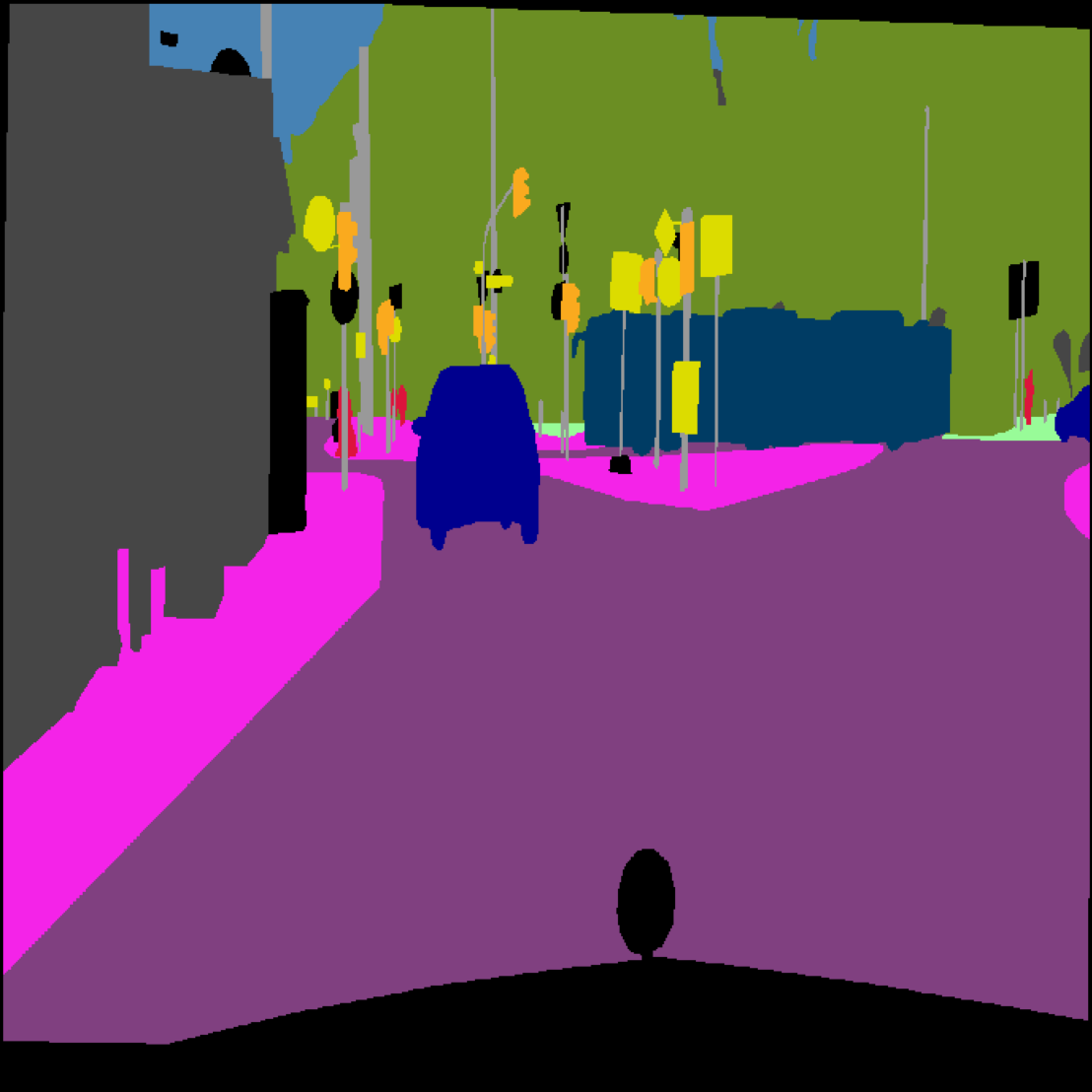}
        \end{center}
    \end{subfigure}
    \hfill
    \begin{subfigure}[t]{0.24\textwidth}
        \begin{center}
        \includegraphics[width=\linewidth]{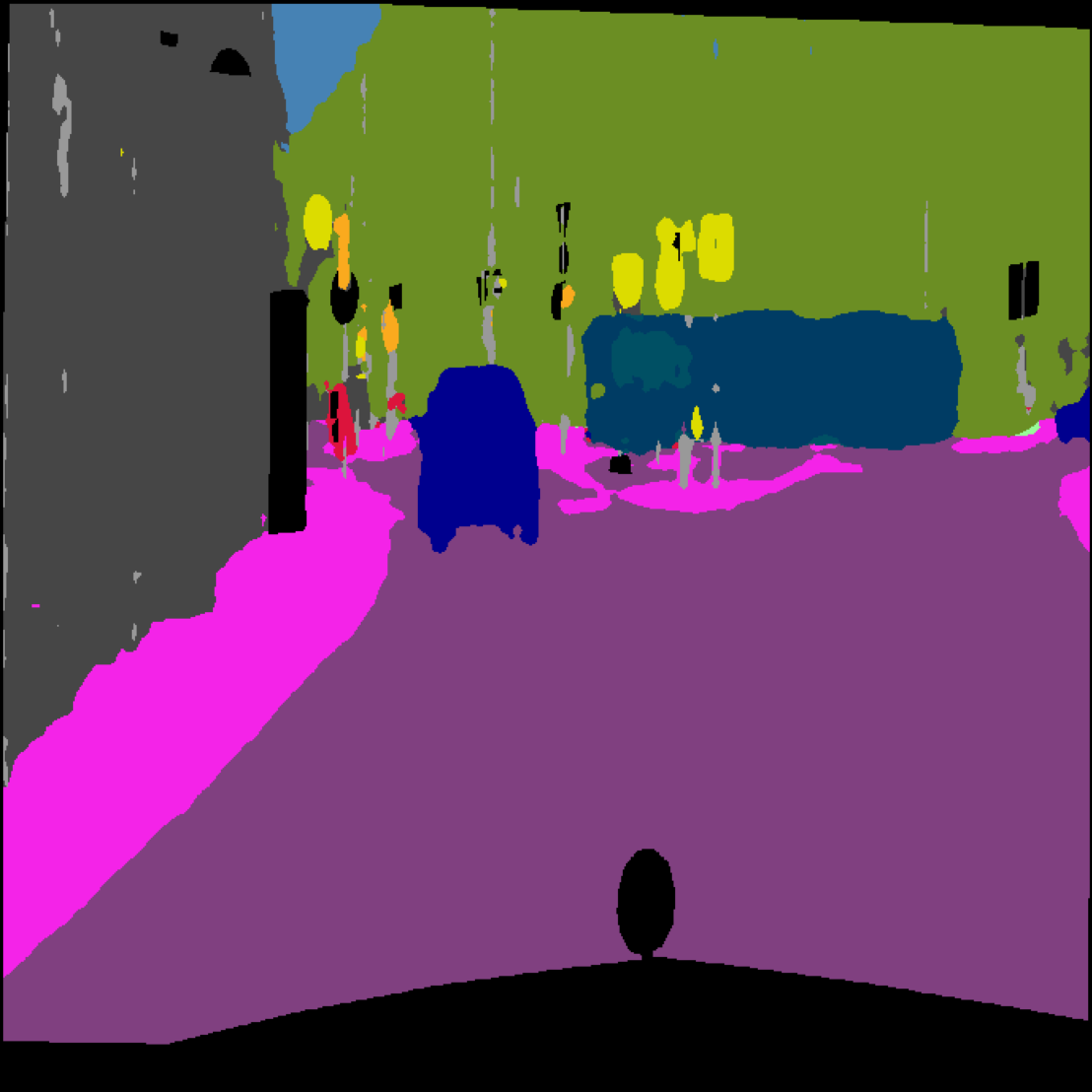}
        \end{center}
    \end{subfigure}
    \hfill
    \begin{subfigure}[t]{0.24\textwidth}
        \begin{center}
        \includegraphics[width=\linewidth]{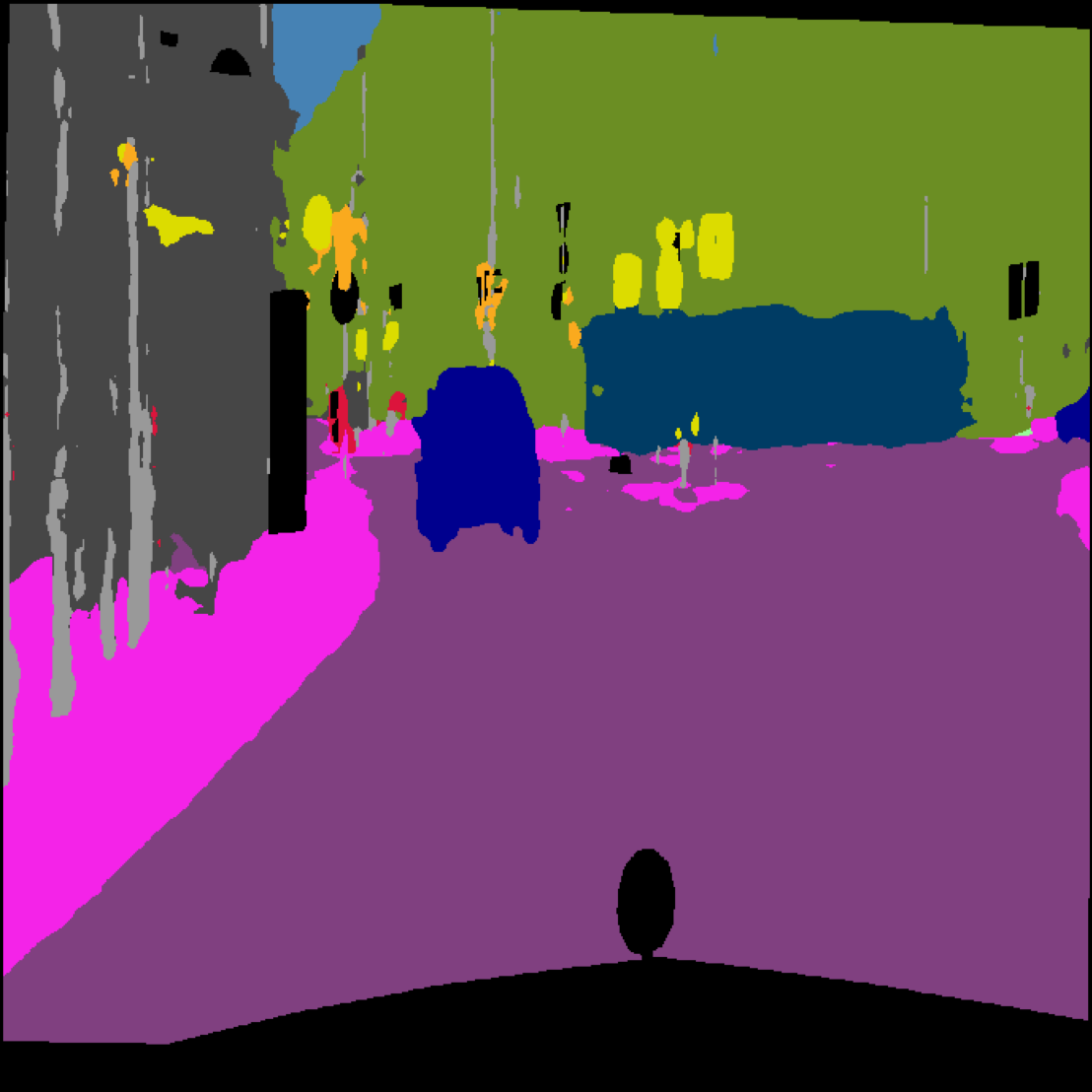}
        \end{center}
    \end{subfigure}
    \vspace*{\fill}
    \begin{subfigure}[t]{0.24\textwidth}
        \begin{center}
        \includegraphics[width=\linewidth]{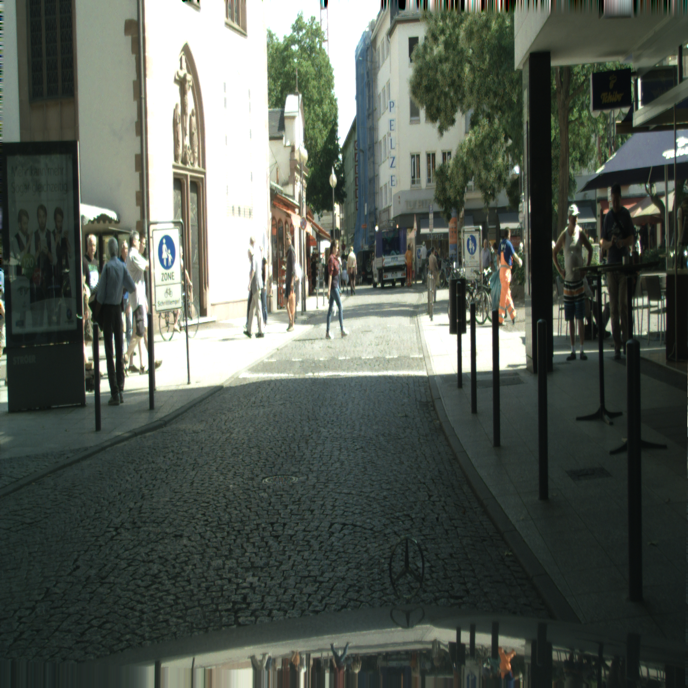}
        \end{center}
    \end{subfigure}
    \hfill
    \begin{subfigure}[t]{0.24\textwidth}
        \begin{center}
        \includegraphics[width=\linewidth]{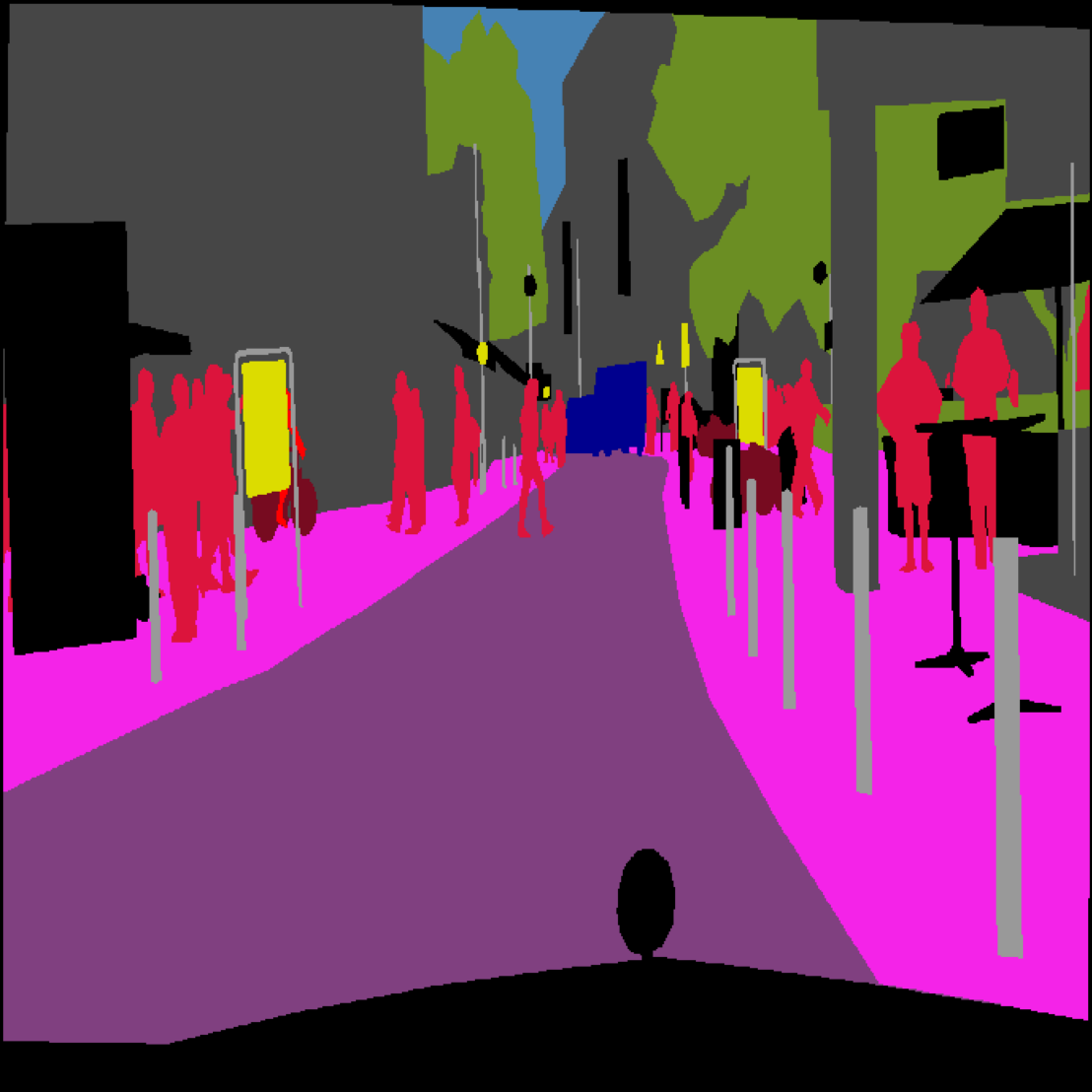}
        \end{center}
    \end{subfigure}
    \hfill
    \begin{subfigure}[t]{0.24\textwidth}
        \begin{center}
        \includegraphics[width=\linewidth]{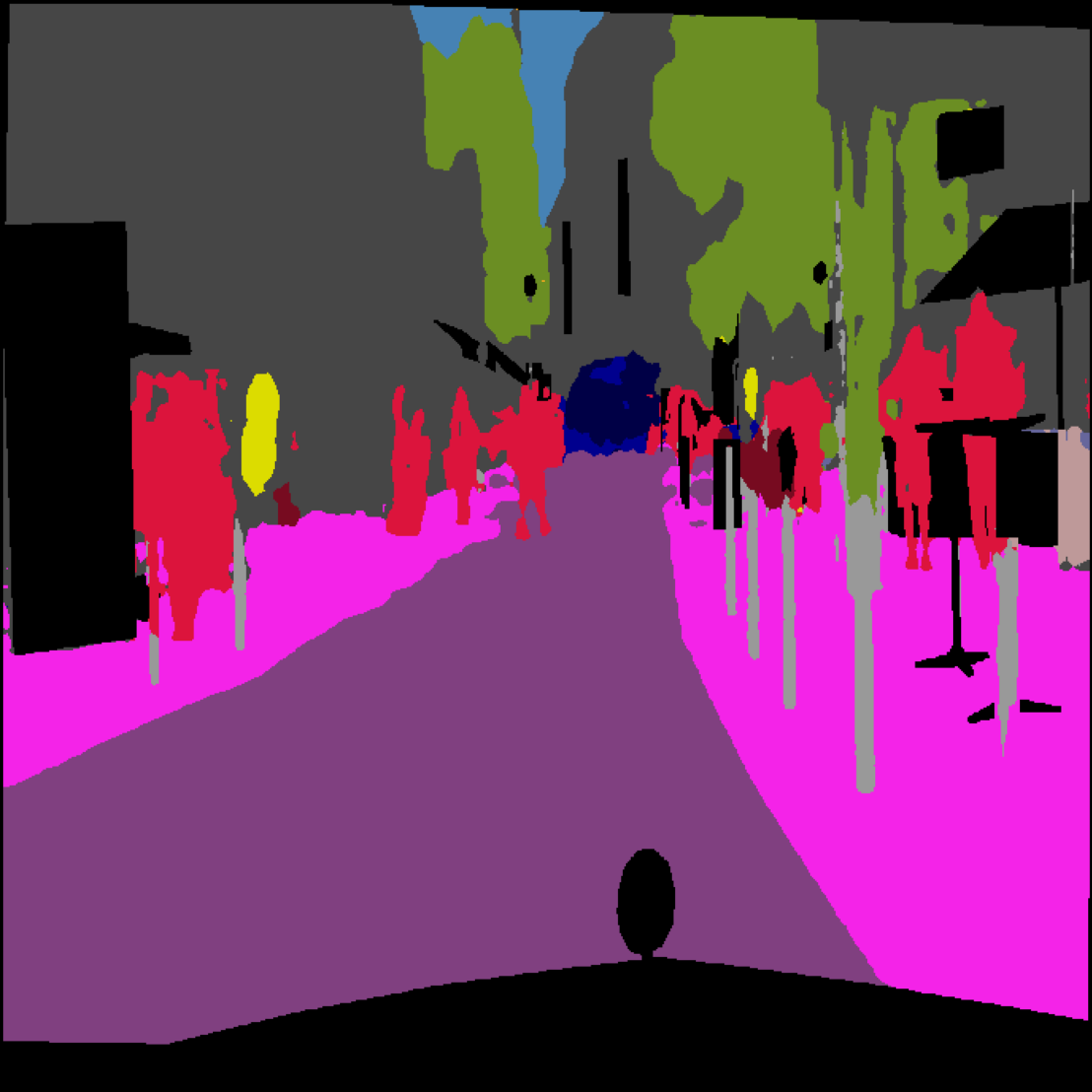}
        \end{center}
    \end{subfigure}
    \hfill
    \begin{subfigure}[t]{0.24\textwidth}
        \begin{center}
        \includegraphics[width=\linewidth]{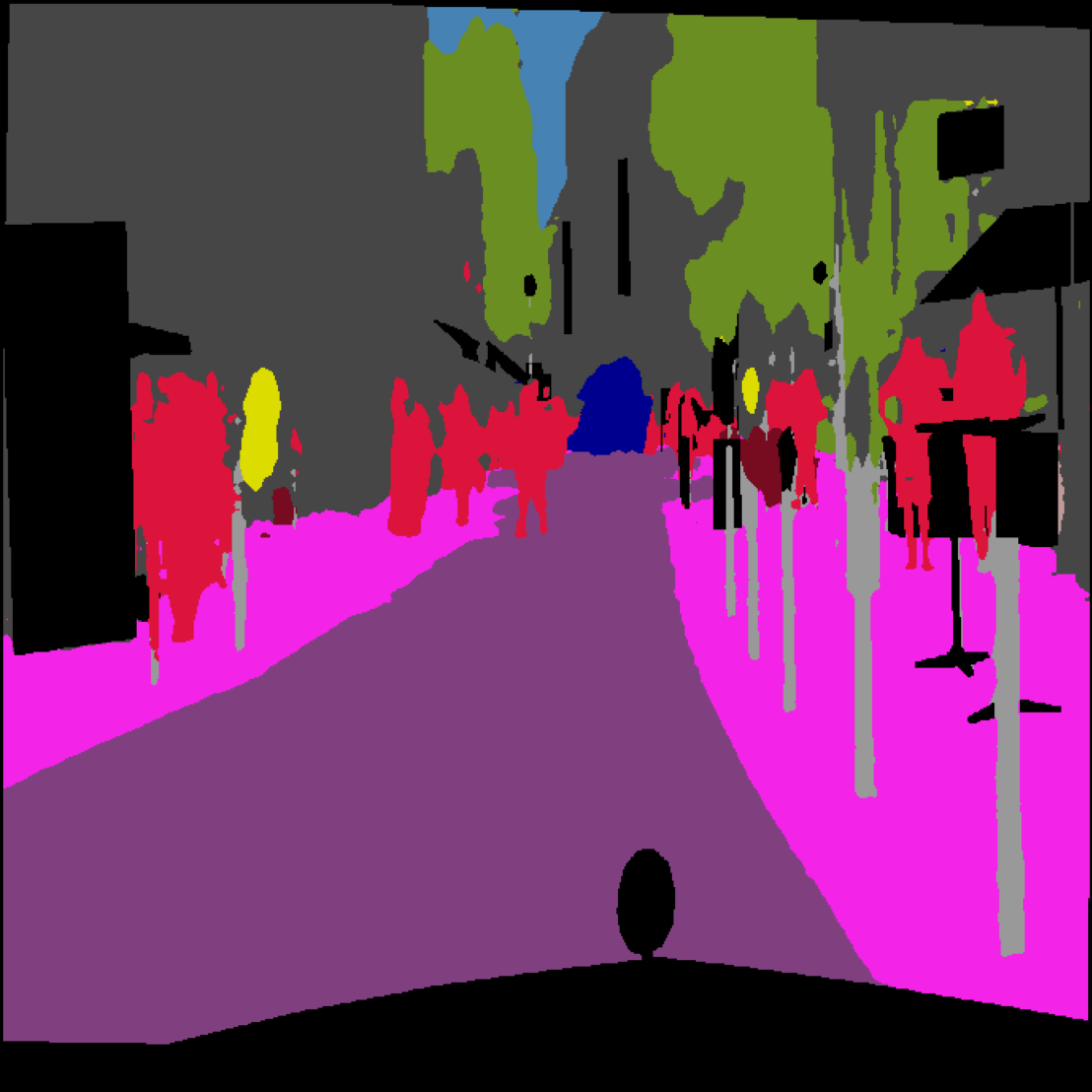}
        \end{center}
    \end{subfigure}
    \vspace*{\fill}
    \begin{subfigure}[t]{0.24\textwidth}
        \begin{center}
        \includegraphics[width=\linewidth]{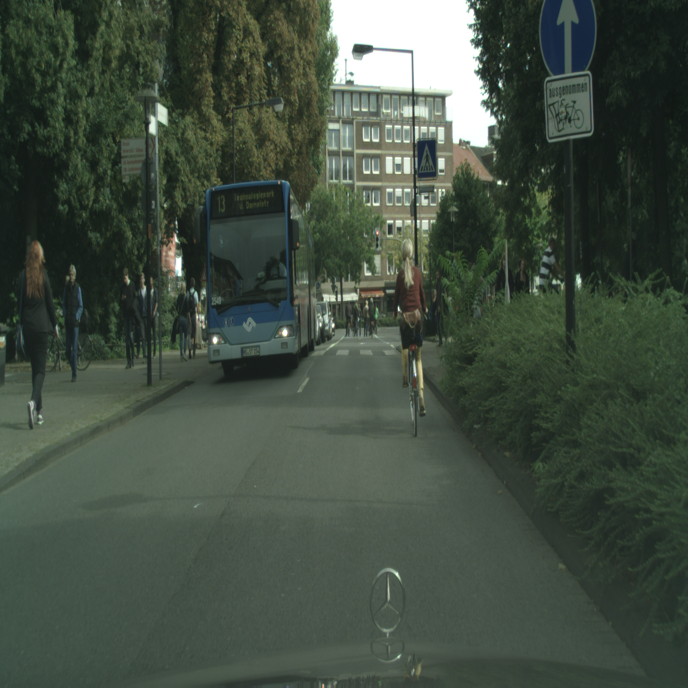}
        \end{center}
    \end{subfigure}
    \hfill
    \begin{subfigure}[t]{0.24\textwidth}
        \begin{center}
        \includegraphics[width=\linewidth]{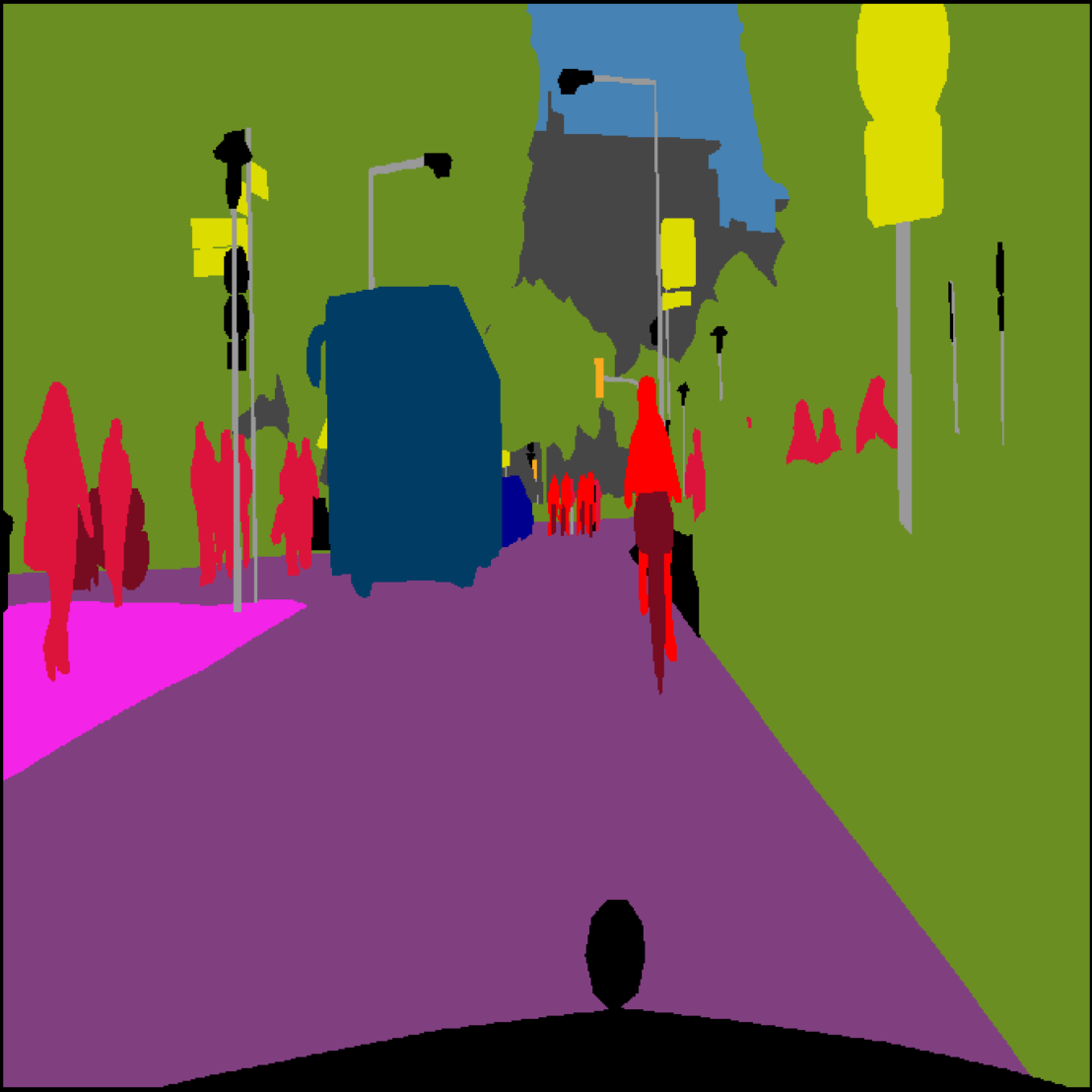}
        \end{center}
    \end{subfigure}
    \hfill
    \begin{subfigure}[t]{0.24\textwidth}
        \begin{center}
        \includegraphics[width=\linewidth]{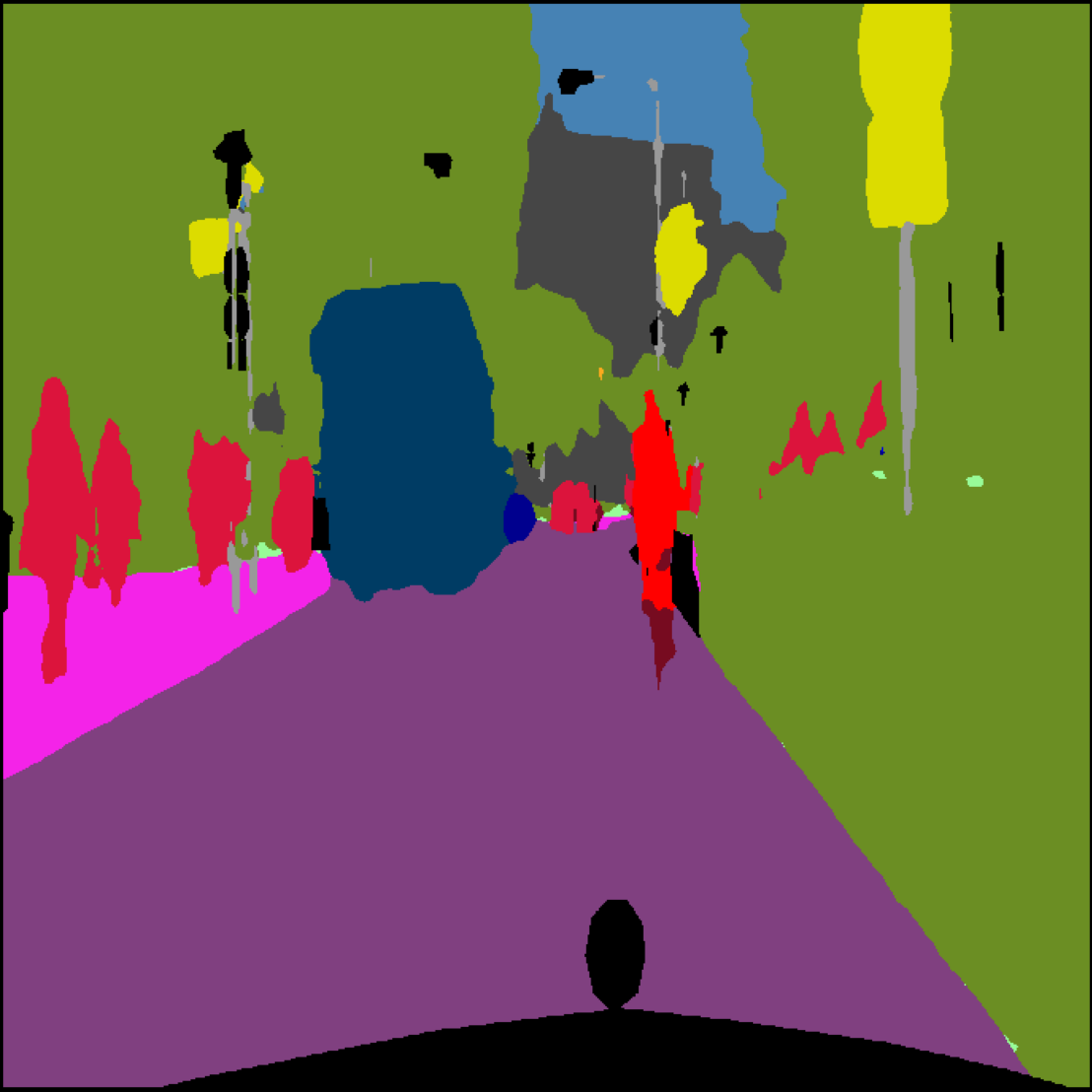}
        \end{center}
    \end{subfigure}
    \hfill
    \begin{subfigure}[t]{0.24\textwidth}
        \begin{center}
        \includegraphics[width=\linewidth]{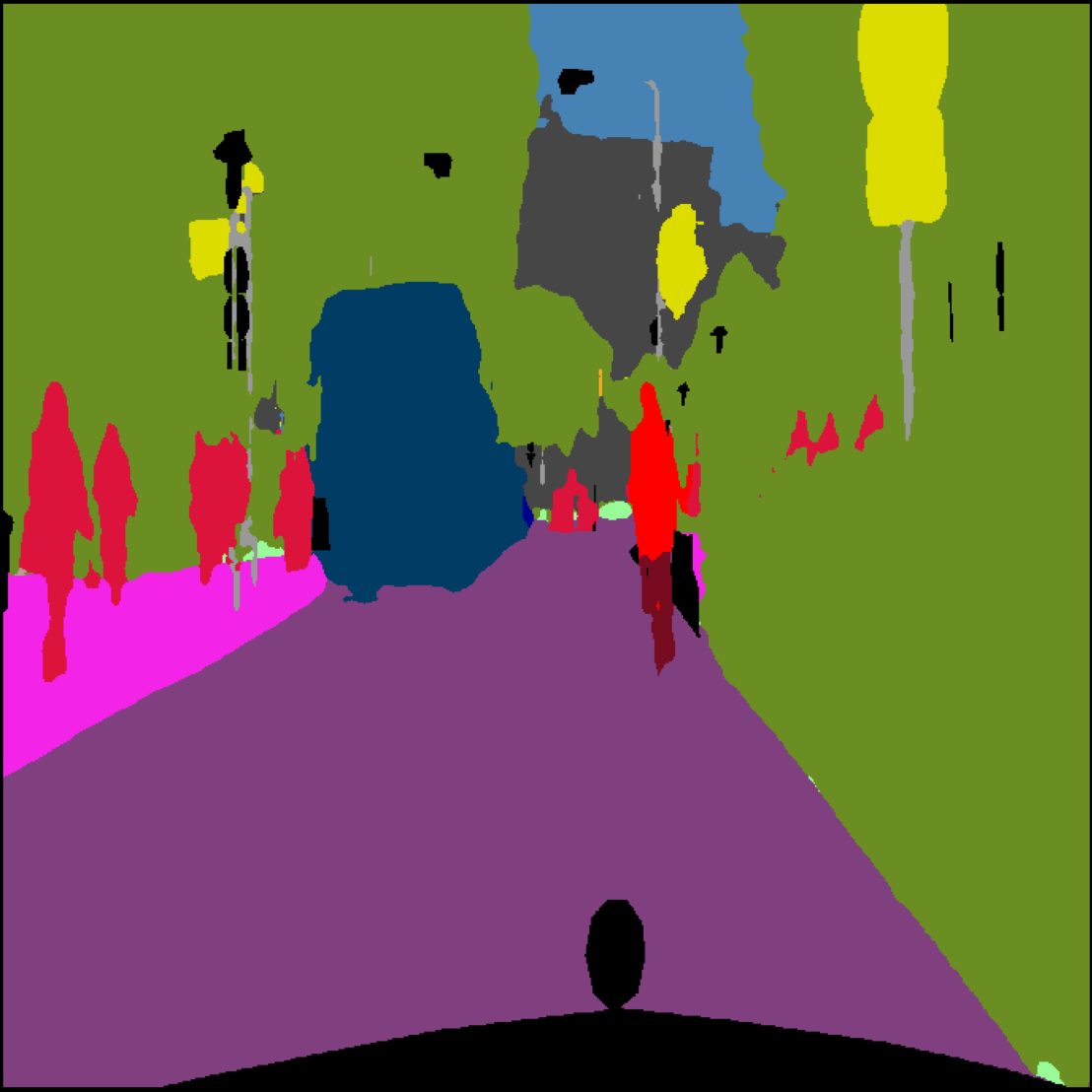}
        \end{center}
    \end{subfigure}
    \vspace*{\fill}
    \begin{subfigure}[t]{0.24\textwidth}
        \begin{center}
        \includegraphics[width=\linewidth]{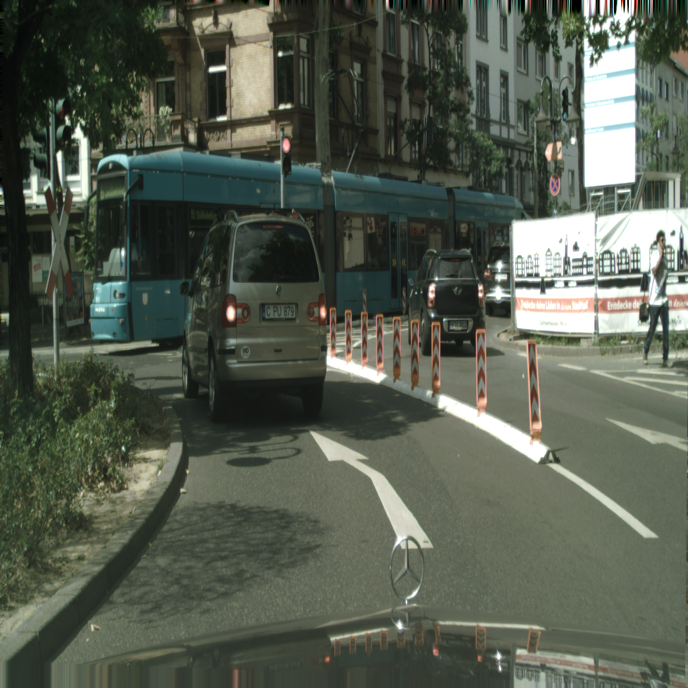}
        \end{center}
    \end{subfigure}
    \hfill
    \begin{subfigure}[t]{0.24\textwidth}
        \begin{center}
        \includegraphics[width=\linewidth]{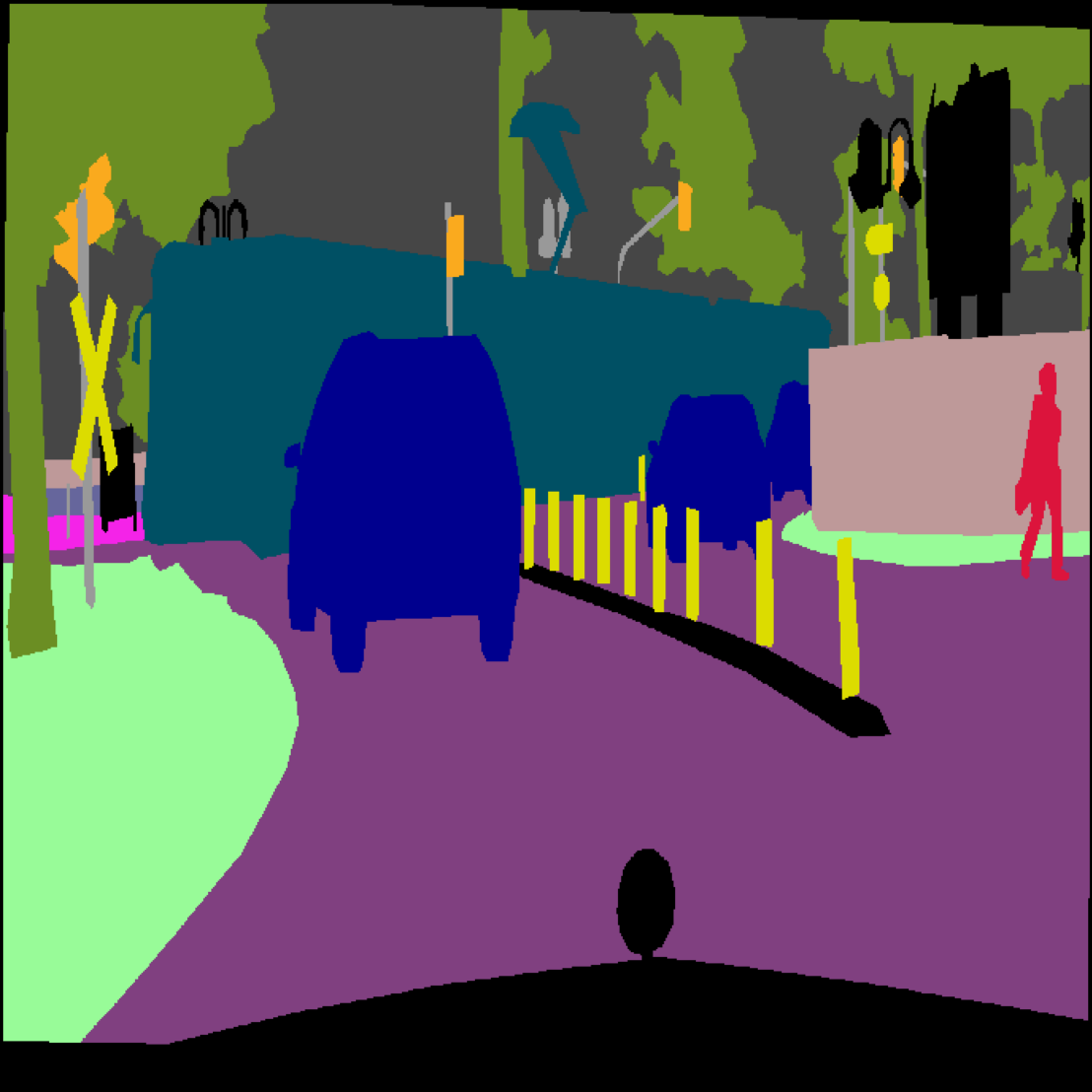}
        \end{center}
    \end{subfigure}
    \hfill
    \begin{subfigure}[t]{0.24\textwidth}
        \begin{center}
        \includegraphics[width=\linewidth]{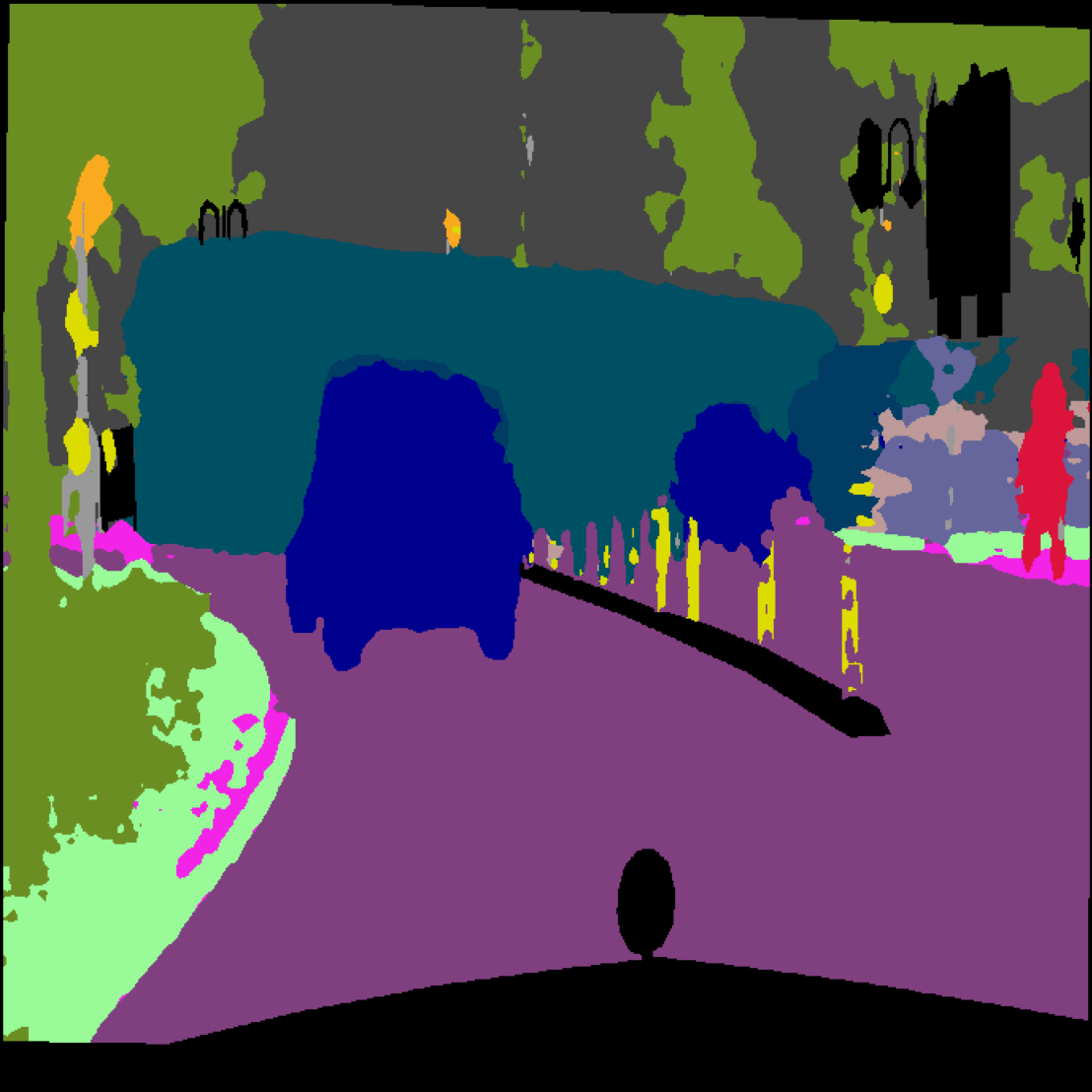}
        \end{center}
    \end{subfigure}
    \hfill
    \begin{subfigure}[t]{0.24\textwidth}
        \begin{center}
        \includegraphics[width=\linewidth]{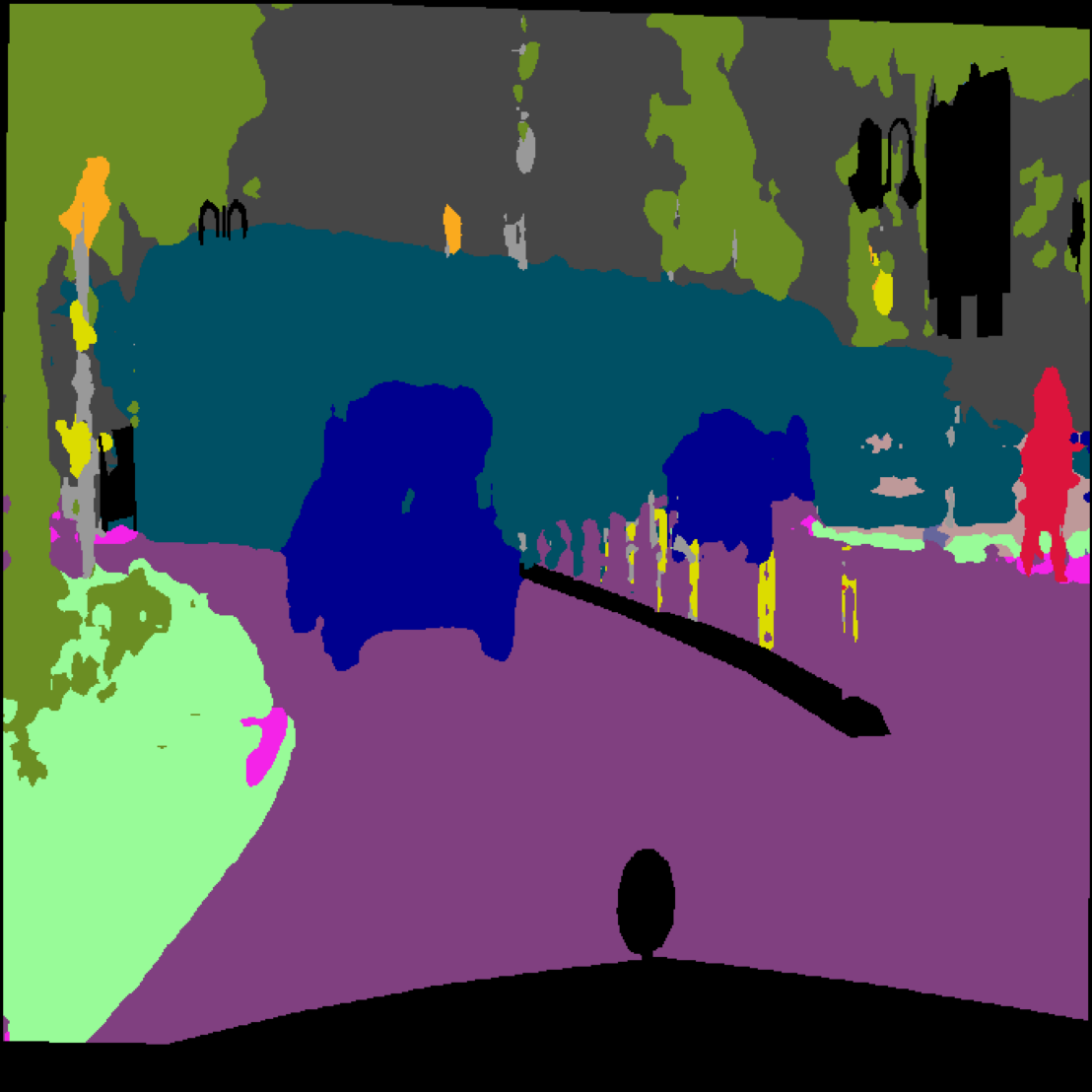}
        \end{center}
    \end{subfigure}
\caption{Examples of semantic segmentation outputs on CityScapes, the columns represent the image (left), the ground truth (center-left), the predictions of a supervised network (center-right), and the predictions of a network trained via S4AL (right).}
\label{fig:viz_dataset_cityscapes}
\end{figure*}

Fig. \ref{fig:selection_dataset_camvid} shows us the important classes belonging to the regions queried for labeling on the CamVid dataset. We can observe high priority is given to the tail distribution classes (in terms of pixel count), namely column pole, pedestrian and bicycle. Regions consisting of areas falling under the ignore category are also sampled, which also proves that the network does not easily assign a particular class from the known categories to a new-ly observed object. Concurrently, Fig. \ref{fig:viz_dataset_camvid} shows us some examples of qualitative results on the CamVid dataset - the first three rows show the network trained with S4AL predicting near similar or better than the fully-supervised network, and the last row shows the most common failure case with respect to the `Fence' class, which tends to get confused with `Building' class due to their structural similarities. 

Similarly, Fig. \ref{fig:viz_dataset_cityscapes} shows us some examples of on the CityScapes dataset - the first three rows show the network trained with S4AL predicting near similar or better than the fully-supervised network, and the last row shows the most common failure case with respect to the `Train' class, which tends to get confused with `Bus' class due to their structural similarities. We observed this in Table \ref{tab:regionvsimage_cityscapes} as well, thus possibly indicating that a hybrid acquisition model of image and regions, on a per image basis, would be most beneficial for complex scenes. Fig. \ref{fig:selection_dataset_cityscapes} shows us that rider, motorcycle and bicycle, three very similar visual categories, were the highest queried regions in the CityScapes dataset. The network also queries a relatively low number of pixels belonging to the `ignore index' label, effectively showing a better sense of understanding for the dataset as most regions for ignoring belong to the hood of the car that is gathering all the data.
\end{document}